\documentclass[twoside,11pt]{article}

\usepackage{blindtext}

\usepackage{jmlr2e}

\usepackage{makecell}
\usepackage{amsmath}
\usepackage{bm}
\usepackage{afterpage}
\usepackage{booktabs}
\usepackage[dvipsnames]{xcolor}
\usepackage{hyperref}
\usepackage{cleveref}
\usepackage{mathrsfs}
\usepackage{float}
\hypersetup{%
  colorlinks=true,
  linkcolor=blue,
  citecolor=blue,
  linkbordercolor=red,
}

\usepackage{lastpage}
\jmlrheading{24}{2023}{1-\pageref{LastPage}}{4/22; Revised
1/23}{1/23}{22-0365}{Shaowu Pan, Steven L. Brunton, and J. Nathan Kutz}
\ShortHeadings{title}{Pan, Brunton, Kutz}

\ShortHeadings{Neural Implicit Flow: a mesh-agnostic dimensionality reduction paradigm}{Pan, Brunton and Kutz}
\firstpageno{1}

\begin{document}

\title{Neural Implicit Flow: a mesh-agnostic dimensionality reduction paradigm of spatio-temporal data}

\author{\name Shaowu Pan \email{shawnpan@uw.edu} \\
       \addr Department of Applied Mathematics\\
       University of Washington\\
       Seattle, WA 98195-4322, USA
       \AND
       \name Steven L. Brunton \email{sbrunton@uw.edu} \\
       \addr Department of Mechanical Engineering\\
       University of Washington\\
       Seattle, WA 98195-4322, USA
       \AND
       \name J. Nathan\ Kutz \email{kutz@uw.edu} \\
       \addr Department of Applied Mathematics\\
       University of Washington\\
       Seattle, WA 98195-4322, USA
       }

\editor{Animashree Anandkumar}

\maketitle

\begin{abstract}
High-dimensional spatio-temporal dynamics can often be encoded in a low-dimensional subspace.  
Engineering applications for modeling, characterization, design, and control of such large-scale systems often rely on dimensionality reduction to make solutions computationally tractable in real time.  
Common existing paradigms for dimensionality reduction include linear methods, such as the {\em singular value decomposition} (SVD), and nonlinear methods, such as variants of {\em convolutional autoencoders} (CAE).
However, these encoding techniques lack the ability to efficiently represent the complexity associated with spatio-temporal data, which often requires variable geometry, non-uniform grid resolution, adaptive meshing, and/or parametric dependencies.
To resolve these practical engineering challenges, we propose a general framework called \textit{Neural Implicit Flow} (NIF) that enables a mesh-agnostic, low-rank representation of large-scale, parametric, spatial-temporal data. 
NIF consists of two modified {\em multilayer perceptrons} (MLPs): (i) {\em ShapeNet}, which isolates and represents the spatial complexity, and (ii) {\em ParameterNet}, which accounts for any other input complexity, including parametric dependencies, time, and sensor measurements. 
We demonstrate the utility of NIF for parametric surrogate modeling, enabling the interpretable representation and compression of complex spatio-temporal dynamics, efficient many-spatial-query tasks, and improved generalization performance for sparse reconstruction.

\end{abstract}



\section{Introduction}

Machine learning and artificial intelligence algorithms have broadly transformed science and engineering, including the application areas of computer vision \citep{krizhevsky2012imagenet}, natural language processing \citep{seq2seq}, molecular dynamics \citep{zhang2018deep,mardt2018vampnets}, and dynamical systems \citep{brunton2019data}. 
The subfield of {\em scientific machine learning}, which often focuses on modeling, characterization, design, and control of large-scale, physics-based models, has also experienced significant growth.
Despite the achievements in scientific machine learning \citep{duraisamy2019turbulence,karniadakis2021physics,kutz2017deep,brunton2020machine}, there remain significant challenges in the representation of high-dimensional spatio-temporal dynamics, which are often modeled by nonlinear partial differential equations (PDEs). 
Data-driven modeling of PDE systems often relies on a more advantageous representation of the physics.
In general, \textit{manifold-based} methods are a dominant paradigm \citep{hesthaven2016certified,carlberg2011efficient,peherstorfer2016data,zahr2015progressive,benner2015survey}.  However, there are recent innovations in developing \textit{mesh-based} methods \citep{long2018pde,zhu2018bayesian,geneva2020modeling,bar2019learning,li2020fourier,pfaff2020learning} and \textit{mesh-agnostic} methods \citep{raissi2020hidden,lu2021learning,lu2021comprehensive,sun2020surrogate}. 
Despite the diversity of algorithmic innovations, each has various challenges in efficiently or accurately representing complex spatio-temporal dynamics.
Indeed, practical engineering applications require handling variable geometry, non-uniform grid resolutions, adaptive meshes, and/or parametric dependencies.
We advocate a general mathematical framework called \textit{Neural Implicit Flow} (NIF) that enables a mesh-agnostic, low-rank representation of large-scale, parametric, spatial-temporal data. 
NIF leverages a hypernetwork structure that allows one to isolate the spatial complexity, thus accounting for all other complexity in a second network where parametric dependencies, time, and sensor measurements are encoded and modulating the spatial layer.  
We show that NIF is highly advantageous for representing spatio-temporal dynamics in comparison with current methods.

Spatio-temporal data is ubiquitous.  As such, a diversity of methods have been developed to characterize the underlying physics.
In manifold-based modeling, which is the most dominant paradigm for reduced-order modeling, one first extracts a low-dimensional manifold from the solution of a PDE, typically using the singular value decomposition~\citep{benner2015survey, noack2003hierarchy, rowley2004model} or a convolutional autoencoder (CAE)~\citep{brunton2019data, holmes2012turbulence,mohan2019compressed,murata2020nonlinear,xu2019multi,lee2019data,ahmed2021nonlinear}.  Then one either directly solves the projected governing equation on the manifold \citep{carlberg2011efficient,carlberg2013gnat} or learns the projected dynamics from the data on the manifold with either Long Short-Term Memory (LSTM)~\citep{mohan2018deep}, Artificial Neural Network~\citep{pan2018long,san2019artificial,lui2019construction}, polynomials \citep{qian2020lift,brunton2016discovering,peherstorfer2016data}. Other variants include jointly learning dynamics together with the manifold \citep{champion2019data,kalia2021learning,lusch2018deep, takeishi2017learning,yeung2019learning, otto2019linearly, pan2020physics, lange2021fourier, mardt2020deep}, and closure modeling to account for non-Markovian effects \citep{pan2018data,wang2020recurrent,maulik2020time,ahmed2021closures}.
%
Manifold-based approaches first reduce the prohibitively large spatial degrees of freedom (e.g., $O(10^4)-O(10^{11})$ in fluid dynamics) into a moderate number (e.g., $O(10)-O(10^2)$) by learning a low-dimensional representation on which we project and solve the PDE, thus inheriting the physics \citep{carlberg2011efficient}, or simply performing time-series modeling on the reduced manifold \citep{xu2019multi} or modal expansion~\citep{taira2017modal}. Equivalently, this preprocessing step can be viewed as learning a time-dependent vector-valued low dimensional representation of a spatio-temporal field.  More broadly, learning an effective low-dimensional representation is often domain specific, for example, using a real-valued matrix with RGB channels for representing images in computer vision \citep{szeliski2010computer}, Word2Vec for representing words in natural language processing (NLP) \citep{mikolov2013efficient}, spectrograms for representing audio signal in speech recognition \citep{flanagan2013speech}, among others. 

Manifold-based methods have several drawbacks when applied to realistic spatio-temporal data.  Specifically, practical engineering applications require handling variable geometry, non-uniform grid resolutions, adaptive meshes, and/or parametric dependencies. This includes data from incompressible flow in the unbounded domain~\citep{yu2022multi}, combustion \citep{bell2011adaptive}, astrophysics \citep{almgren2010castro}, multiphase flows \citep{sussman2005parallelized}, and fluid-structure interactions \citep{bhalla2013unified}) which are generated with advanced meshing techniques such as {\em adaptive mesh refinement} (AMR) \citep{berger1984adaptive}, and/or overset grids \citep{chan2009overset}. Such meshes typically change with time or parameters (e.g., Mach number dependency on shock location) in order to efficiently capture the multi-scale phenomena\footnote{for some systems \citep{bryan2014enzo}, AMR is the only way to make such simulation possible.}, which violates the requirement of common SVD approaches. While CNNs require preprocessing the flowfield as an image, i.e., voxel representation with a uniform Cartesian grid, in order to perform discrete convolution, which is affordable in 2D but becomes increasingly expensive in 3D \citep{park2019deepsdf} due to cubic memory requirements. The memory footprint limits the resolution to $64^3$ typically \citep{mescheder2019occupancy} unless one resorts to an optimized parallel implementation of 3D CNN on clusters \citep{mathuriya2018cosmoflow}. Additionally, such \textit{uniform} processing is inherently inconsistent with the \textit{nonuniform} multi-scale nature of PDEs and can decrease the prediction accuracy of downstream tasks, e.g., failing to accurately recover total lift/drag from flowfield predictions \citep{bhatnagar2019prediction} or efficiently capture small-scale geometry variations that can trigger critical physics phenomena. This leads us to pursue a mesh-agnostic and expressive paradigm beyond SVD and CAE for dimensionality reduction of spatio-temporal data.


Recent advances in mesh-based methods\footnote{\textcolor{black}{In this paper, ``mesh-based'' in our paper is the same as ``graph-based''. A key characteristics of graph-based method is that the online computational cost scales with the size of the graph (i.e., how much grid points) \citep{chen2021model}. While manifold-based methods only access mesh during postprocessing. }} have shown promising results either with discretization-invariant operator learning~\citep{li2020fourier,li2020multipole,li2020neural} or meshes based on the numerical solver~\citep{pfaff2020learning,sanchez2020learning,xu2021conditionally}. On the other hand, a mesh-agnostic framework (e.g., physics-informed neural network (PINN) \citep{raissi2020hidden}) has been successfully applied to solving canonical PDEs on problems where mesh-based algorithms can be cumbersome due to, for instance, arbitrary geometry \citep{berg2018unified,sun2020surrogate} and/or high-dimensionality \citep{sirignano2018dgm}. In addition to leveraging the PDE governing equation, it successfully leverages the structure of multilayer perceptrons (MLPs) with the coordinate as input and the solution field as output. It can be viewed as using an adaptive global basis in space-time to approximate the solution with the known PDE, instead of the traditional local polynomial basis inside a cell, e.g., finite element analysis \citep{hughes2012finite}. Concurrently, such MLP structure has been also employed in computer graphics community for learning 3D shape representations \citep{park2019deepsdf,mescheder2019occupancy}, scenes \citep{mildenhall2020nerf,sitzmann2020implicit,tancik2020fourier}. 
\textcolor{black}{A closely related work called MeshfreeFlowNet (MFN) \citep{esmaeilzadeh2020meshfreeflownet} uses CNN and coordinate-based MLP to perform super-resolution for fluid problems. MFN first uses a 2D CNN to extract features from coarse-scale spatial-temporal field. Then, the extracted features are concatenated with $t,x,y$ as an augmented input to a MLP, which outputs the high-resolution measurement at $t,x,y$. }

Motivated by the above seminal works, we introduce a mesh-agnostic representation learning paradigm called \textit{neural implicit flow} (NIF) that exploits the expressiveness and flexibility of the multilayer perceptrons (MLPs) for dimensionality reduction of parametric spatio-temporal fields. In \cref{sec:framework}, we present the NIF paradigm and several derived problem-specific frameworks. In \cref{sec:applications}, we demonstrate the following capabilities of NIF: 
\begin{enumerate}
\item NIF enables an efficient scalable 3D nonlinear dimensionality reduction on spatial-temporal datasets from arbitrary different meshes. Example includes a three-dimensional fully turbulent flows with over 2 million cells. (see \cref{sec:app_application_hit})

\item NIF also enables modal analysis of spatial-temporal dynamics on adaptive meshes. As an example, we explore dynamic mode decomposition on the fluid flow past a cylinder with adaptive mesh refinement (see \cref{sec:app_modal_analysis}).

\end{enumerate}

\noindent NIF also provides a performance improvement in the following applications on several canonical spatio-temporal dynamics: 
\begin{enumerate}
\item NIF generalizes 40\% better in terms of root-mean-square error (RMSE) than a generic mesh-agnostic MLP in terms of mesh-agnostic surrogate modeling for the parametric Kuramoto–Sivashinsky PDE under the same size of training data or trainable parameters (see \cref{sec:app_data-fit-1d-ks})

\item NIF outperforms both (linear) SVD and (nonlinear) CAE in terms of nonlinear dimensionality reduction, as demonstrated on the Rayleigh-Taylor instability on adaptive mesh with a factor from 10 times to 50\% error reduction. (see \cref{sec:app_better_compression})

\item Compared with the original implicit neural representation which takes \emph{all} information (including time $t$) into a single feedforward network~\citep{sitzmann2020implicit,lu2021compressive}, NIF enables efficient spatial sampling with 30\% less CPU time and around 26\% less memory consumption under the same level of accuracy for learning the aforementioned turbulence dataset.
(see \cref{sec:app_efficient_spatial_query})

\item NIF outperforms the state-of-the-art method (POD-QDEIM~\citep{drmac2016new}) in the task of data-driven sparse sensing with 34\% smaller testing error on the sea surface temperature dataset. (see \cref{sec:app_sparse_sensing})

\end{enumerate}

\noindent
Finally, conclusions and future perspectives of NIF are presented in \cref{sec:conclusion}.

\section{Neural Implicit Flow}
\label{sec:framework}

We begin by considering 3D spatial-temporal data with varying time/parameters, namely $\mathbf{u}(\mathbf{x}; t, \bm{\mu}) \in \mathbb{R}^{n}$,
with spatial coordinate $\mathbf{x} \in \mathbb{R}^3$, time $t \in  \mathbb{R}^{+}$, and parameters $\bm{\mu} \in  \mathbb{R}^{d}$ (e.g., Reynolds number).  Without loss of generality, consider a supervised learning problem: using an $L$-layer MLP with spatial coordinate $\mathbf{x}$ as input to fit a \textit{single} spatial realization at an arbitrary time $t_0$ and parameter $\bm{\mu}_0$, i.e.,  $\mathbf{u}(\mathbf{x}; t_0, \bm{\mu}_0)$.
An $L$-layer MLP with $\mathbf{x}$ as input is a vector-valued function defined as $\mathbf{u}_{\textrm{MLP}}(\mathbf{x}; \mathscr{W}, \mathscr{B}) = \mathbf{W}_L \eta_L + \mathbf{b}_L$, $\forall l \in \{1,\ldots,L-1\}$ with $\eta_l = \sigma (\mathbf{W}_l \eta_{l-1} + \mathbf{b}_l)$. 
The first layer weight $\mathbf{W}_1$ has size $n_h \times 3$, and the remaining hidden layer weight $\mathbf{W}_l$ has size $n_h \times n_h$. The last layer weight $\mathbf{W}_L$ has size $n \times n_h$. Biases are denoted as $b_l$ where subscript $l$ denotes the index of layer.
%
$\eta_0 = \mathbf{x}$. The set of weights and biases are defined as $\mathscr{W}=\{\mathbf{W}_l\}_{l=1}^{L}$ and $\mathscr{B}=\{\mathbf{b}_l\}_{l=1}^{L}$ correspondingly. The activation $\sigma: \mathbb{R} \mapsto \mathbb{R}$ is a non-linear continuous function. One can use gradient-based optimization to find the weights $\mathscr{W}$ and biases $\mathscr{B}$ that minimize the discrepancy between $\mathbf{u}_{\textrm{MLP}}$ and the single snapshot data $\mathbf{u}$ at $t$ and $\bm{\mu}$: 
\begin{equation}
\label{eq:discrepy_single_snap}
    \min_{\mathscr{W}, \mathscr{B}} \int \mathcal{L} (\mathbf{u}_{\textrm{MLP}}(\mathbf{x}; \mathscr{W}, \mathscr{B}), 
    \mathbf{u}(\mathbf{x}; t_0, \bm{\mu}_0)) d\nu (\mathbf{x}; t_0, \bm{\mu}_0),
\end{equation}
where $\mathcal{L}$ is a loss function (e.g., mean squared error, mean absolute error) and $\nu$ is some spatial measure which can depend on $t_0, \bm{\mu}_0$. 
This minimization leads to a key observation: a well-trained set of weights $\mathscr{W}$ and biases $\mathscr{B}$ of the MLP fully determines a spatial field that approximates the target field, in a mesh-agnostic sense. Such a trained MLP (what we call \textit{ShapeNet}) is closely related to the so-called \textit{neural implicit} representation \citep{sitzmann2020implicit}, and works in computer graphics \citep{park2019deepsdf,mescheder2019occupancy} used such an MLP to fit signed distance functions (SDFs) of a desired surface, which is \textit{implicitly} defined by the zero iso-surface of the SDF. 

\begin{figure}[t]
\centering
\includegraphics[width=1\textwidth]{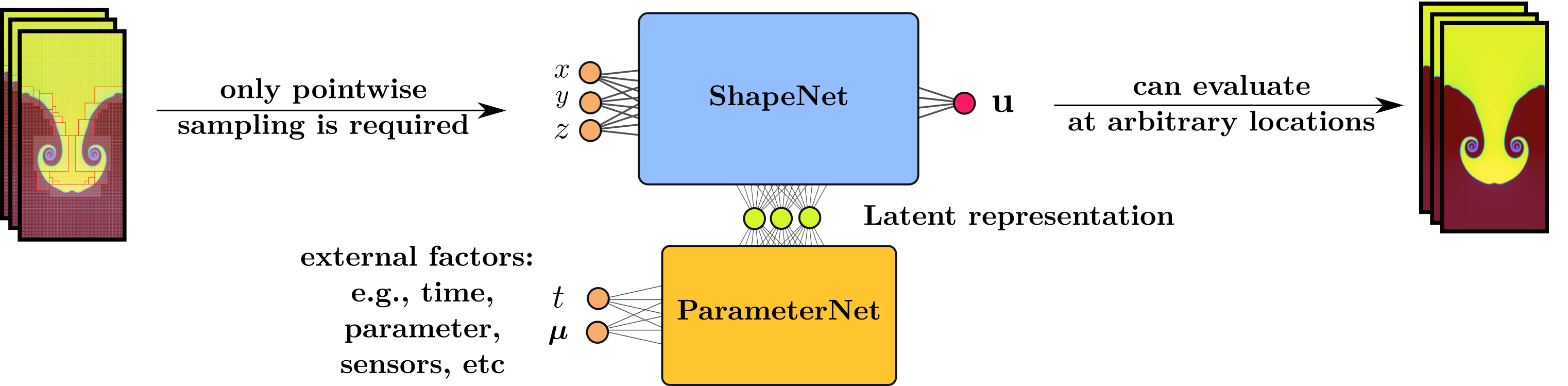}
\caption{Neural implicit flow framework for dimensionality reduction of spatio-temporal field from PDE. NIF uses the bottleneck layer, which is linearly mapped to weights $\mathscr{W}$ (and biases $\mathscr{B}$) of ShapeNet, as the latent representation. ParameterNet correlates external factors with such a latent representation.}
\label{fig1}
\end{figure}

To explicitly correlate the corresponding spatial fields with external factors of interests, such as time, parameters, and sparse sensor measurements, we use a second MLP (denoted as \textit{ParameterNet}) to learn mappings from these parameters to weights $\mathscr{W}$ and biases $\mathscr{B}$. Note that the typical output dimension of ParameterNet, e.g.,, the total number of scalars in $\mathscr{W}$ and $\mathscr{B}$, is above thousands to tens of thousands. 
Dimensionality reduction assumes the existence of a rank-$r$ subspace that can approximate the spatio-temporal data. The reduced coordinates are denoted as $\zeta_1,\ldots,\zeta_r$. One can simply consider a \emph{linear} output layer in ParameterNet after a bottleneck layer of width $r$. Once $\zeta_1,\ldots,\zeta_r$ are determined, $\mathscr{W}$ and $\mathscr{B}$  in the ShapeNet are completely determined, which in turn determines the spatial  \textit{``flow''} field conditioned on the time $t$ or more generally, any other external parameters. 
As a result, the bottleneck layer in \cref{fig1} can be viewed as an $r$-dimensional latent representation, similar to the bottleneck layer in generic autoencoders \citep{goodfellow2016deep}\footnote{However, note that we haven't introduced ``encoder'' in \cref{fig1}. We will introduce an ``encoder'' based on sparse sensors in \cref{sec:app_better_compression}.}. 
In summary, we can find a mesh-agnostic representation of parametric spatio-temporal fields as in the aforementioned manifold-based methods, where spatial complexity, e.g., coherent structures, is explicitly decoupled from temporal and parametric complexity, e.g., chaos and bifurcations.  

Note that any neural network that generates the weights and biases of another neural network is generally called a \textit{hypernetwork} \citep{ha2016hypernetworks}. This concept has been applied in image \citep{sitzmann2020implicit} and language modeling \citep{ha2016hypernetworks} and its inspiration can be dated back to control fast-weight memory \citep{schmidhuber1992learning} in the early 90's. Therefore, NIF can be viewed as a special family of hypernetworks for MLP with only spatial input $\mathbf{x}$ while any other factors, e.g., time $t$, parameter $\bm{\mu}$, are fed into the hypernetwork. This implies the dimensionality reduction is only performed towards spatial complexity, which is the biggest cause of computational and storage challenges for non-linear PDEs. Thus, spatial complexity is decoupled from other factors. 

\begin{figure}[t]
\centering
\includegraphics[width=1\textwidth]{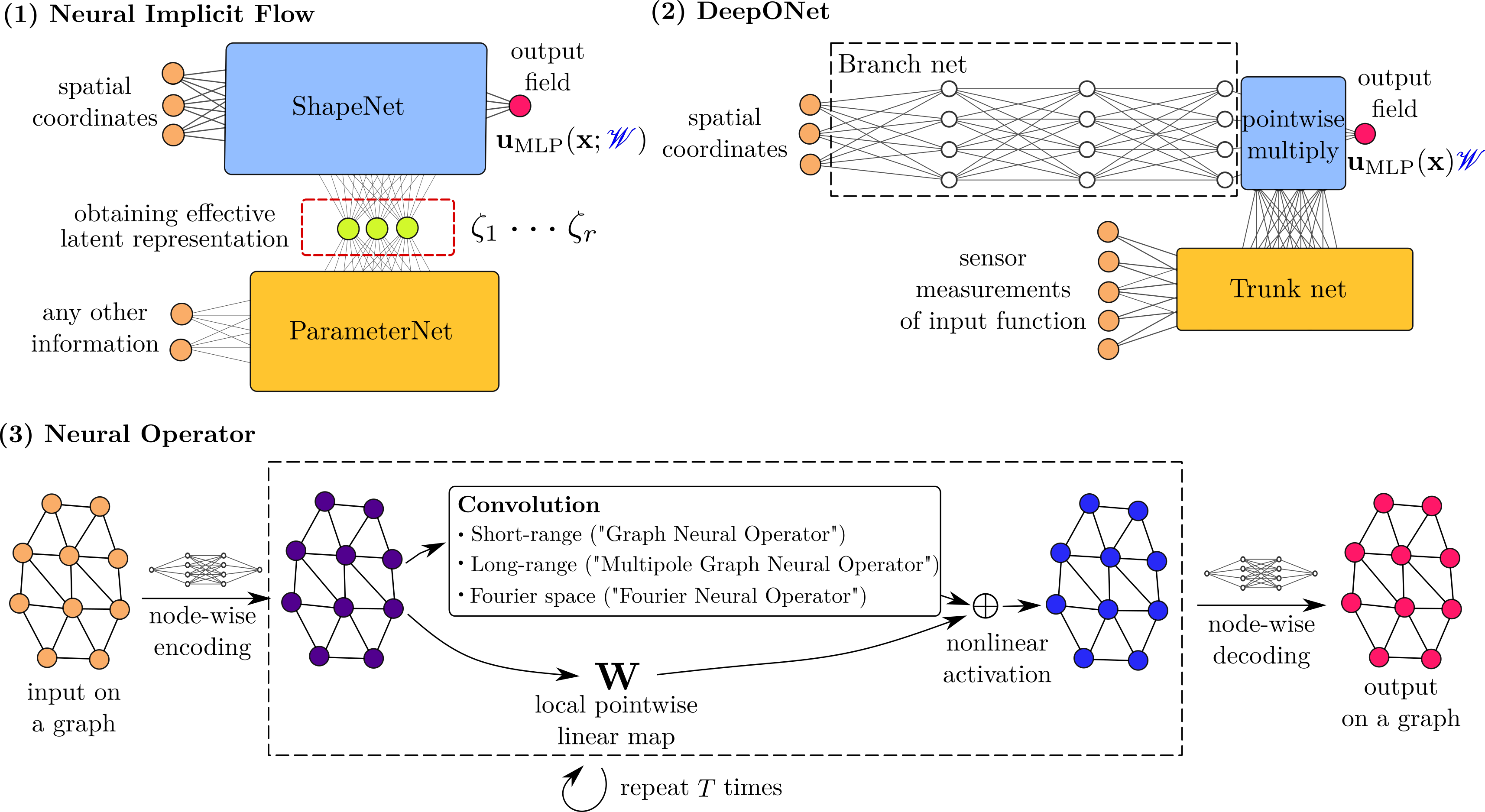}
\caption{Comparion of NIF with DeepONet~\citep{lu2021learning} and Neural Operator~\citep{li2020neural,li2020multipole,li2020fourier}. Notice that the skeleton of DeepONet can be viewed as that of a \emph{last layer parameterized} NIF. Information from sensor measurements is parameterized as a matrix multiplication while NIF is parameterized everywhere inside the ShapeNet. Neural Operator learns continuous nonlinear convolution in order to be mesh-invariant and spatial invariant. Neural Operator and DeepONet focus on learning solution operator while NIF focus on learning a reduced latent representation of spatial-temporal dynamics on arbitrary mesh.}
\label{fig:nif-compare-with-others}
\end{figure}

\begin{figure}[t]
\centering
\includegraphics[width=1\textwidth]{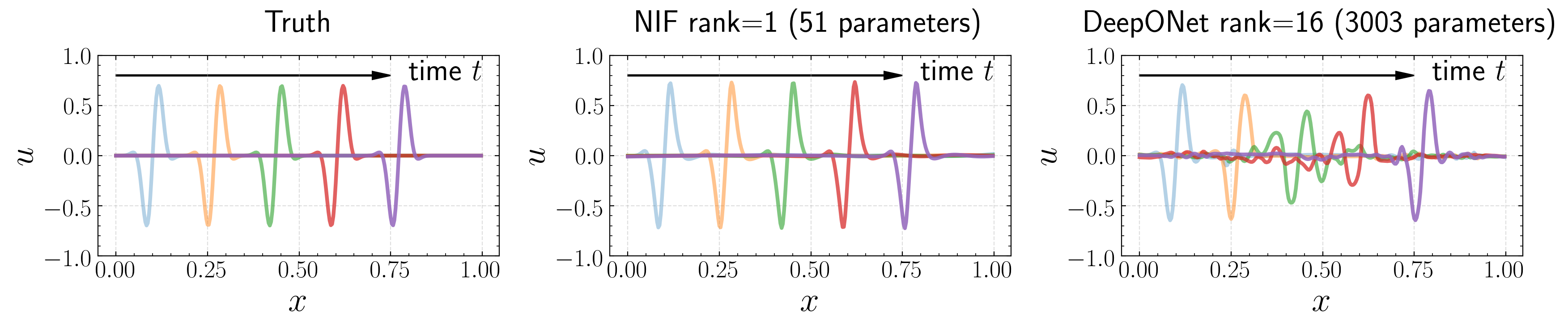}
\caption{Comparison between DeepONet and NIF on a 1D modulated traveling sine wave traveling from left to right. The wave is described as $u(x,t)=e^{-c_0(x-x_0-ct)^2}\sin(\omega(x-x_0-ct))$ with $c_0=-1000,c=0.012,\omega=70$. Data is generated by uniformly sampling $x$ in $[0,1]$ with 300 samples and $t$ in $[0,70]$ with 20 samples. Data is then standard normalized for training. Configuration of NIF: ShapeNet with input $x$: 1-2-2-2-1, ParameterNet with input $t$: 2-2-1-19. Configuration of DeepONet: branch net with input $t$: 1-30-30-17. trunk net with input $x$: 1-30-30-16-1.}
\label{fig:nif-deeponet-travelling-wave}
\end{figure}

A comparison of the NIF architecture with other recent frameworks of machine learning for PDEs is shown in \cref{fig:nif-compare-with-others}. DeepONet~\citep{lu2021learning,wang2021learning} pioneered the learning of general operators associated with PDEs from data. It has been a building block of a DeepM{\&}MNet~\citep{cai2021deepm}, which has been applied to data assimilation in electroconvection \citep{cai2021deepm} and hypersonic flows~\citep{mao2021deepm}. Interestingly, the structure of DeepONet can be viewed as that of NIF with only the last linear layer of the ShapeNet determined by the so-called trunk net.
To highlight the expressiveness of NIF, we compare a tiny NIF with only 51 parameters and a moderate DeepONet with 3003 parameters on a 1D modulated traveling sine wave. \Cref{fig:nif-deeponet-travelling-wave} shows no visual difference between the ground truth and NIF. 
%

Alternatively, neural operators~\citep{li2020neural,li2020multipole,li2020fourier,kovachki2021neural} are discretization-independent mesh-based frameworks for learning solution operators of PDEs beyond the straightforward CNN approach on a fixed grid~\citep{zhu2018bayesian,long2018pde}. Neural operators have been successful in recovering Green's function, learning chaotic PDE systems~\citep{li2021markov}, and leveraging known PDEs such as Navier-Stokes equations~\citep{li2021physics}. It inherits translational and rotational invariance using graph neural networks, as opposed to NIF and DeepONet, which might be advantageous in the small data regime. 

Distinct from the above pioneering works aimed at learning solution operators of PDEs, NIF emphasizes a scalable nonlinear dimensionality reduction paradigm that outperforms SVD and CAE in dealing with complex parametric spatial-temporal data (e.g., turbulence) on \textit{arbitrary mesh structure}. In the following subsections, we will develop problem-specific, NIF-based frameworks on a variety of learning tasks.


\subsection{Data-fit parametric surrogate modeling for PDEs}
\label{sec:datafit}

\begin{figure}[t]
\centering
\includegraphics[width=0.95\textwidth]{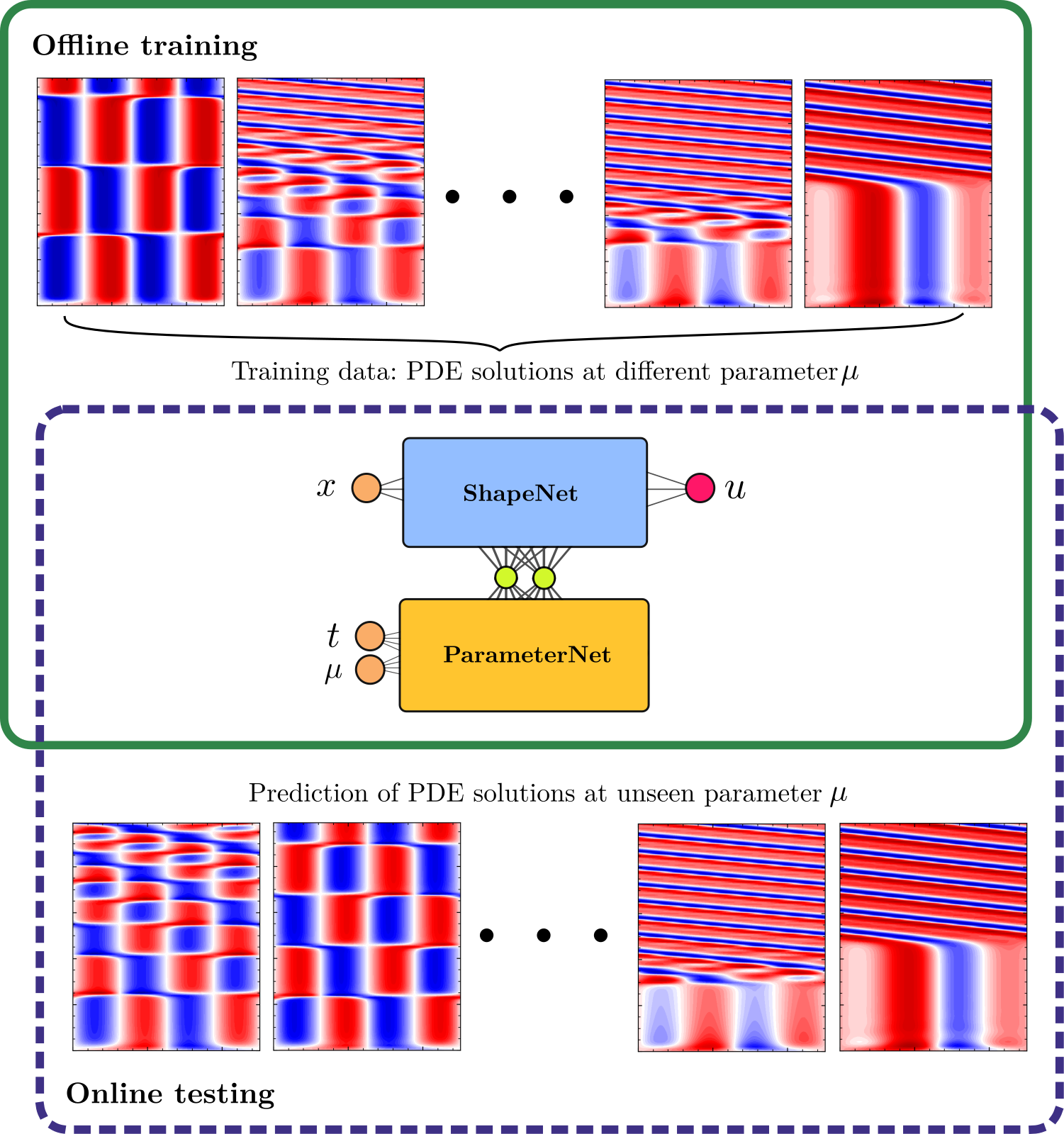}
\caption{Application of NIF on mesh-agnostic surrogate modeling of parametric PDE. The case of 1D parametric Kuramoto-Sivashinsky equation is taken for illustration and further studied in \cref{sec:app_data-fit-1d-ks}. }
\label{fig-r1}
\end{figure}

Consider a class of non-linear parametric PDEs, ${\partial \mathbf{u}}/{\partial t} = \mathcal{G}(\bm{\mu}, \mathbf{u}, \nabla \mathbf{u}, \nabla^2 \mathbf{u}, \ldots)$, $(\mathbf{x}, t, \bm{\mu}) \in \Omega = \mathcal{X} \times \mathcal{T} \times \mathcal{D}$. Here $\mathcal{X} \subset \mathbb{R}^3, \mathcal{T} \subset \mathbb{R}^{+}, \mathcal{D} \subset \mathbb{R}^d$ and $\mathcal{G}$ is a non-linear function or operator in general.  An example of data-fit parametric surrogate modeling \citep{benner2015survey,sobieszczanski1997multidisciplinary,frangos2010surrogate,amsallem2013model,bhatnagar2019prediction} is to find an approximated relation between the parameter $\bm{\mu}$ and the corresponding PDE solution $\mathbf{u}(\mathbf{x}; t, \bm{\mu})$ under fixed initial and boundary conditions. After the data-fit surrogate model is trained, it is expected to make predictions for an unseen input possibly in \textit{real-time}\footnote{ \textcolor{black}{Note that our major goal in the following PDE examples is to demonstrate NIF for dimensionality reduction, not for learning solution operator of PDEs as is advocated in other neural network architectures. However, we present the following example just to show the capability of efficient surrogate modeling where computational cost is independent of mesh size. Note that existing graph-based frameworks still suffer from poor scaling of computational cost with mesh size.}. }. An illustration of the method is shown in \cref{fig-r1}. This is attractive for many engineering tasks that require \textit{many-query} analyses such as optimization, uncertainty quantification, and control. In contrast to more physics-based models \citep{benner2015survey}, a data-fit model simplifies the surrogate modeling for PDEs as a high-dimensional regression without access to prior knowledge. Despite the many disadvantages, including large sample complexity, lack of interpretability etc., it is the simplest and most widely used type of surrogate model~\citep{qian2020lift,loiseau2021pod}.

As illustrated in \cref{fig-r1}, we apply NIF to the above parametric surrogate modeling by simply allowing the weights $\mathscr{W}$ and biases $\mathscr{B}$ to depend on time and parameters through the ParameterNet $f_{\textrm{MLP}}$, 
\begin{equation}
    \label{eq:data-fit}
    \begin{bmatrix}
    \textrm{vec}^\top\left(\mathbf{W}_1 \right)&  \ldots & \textrm{vec}^\top\left(\mathbf{W}_L\right) &
    \mathbf{b}^\top_1 & \ldots & \mathbf{b}^\top_L
    \end{bmatrix} = f_{\textrm{MLP}}(t, \bm{\mu}; \mathbf{\Theta}),
\end{equation}
where $f_{\textrm{MLP}}: \mathbb{R}^{+} \times \mathbb{R}^d \mapsto \mathbb{R}^{m}$ is an MLP with its own weights and biases denoted as $\mathbf{\Theta}$, $\textrm{vec}$ is the matrix vectorization, and $m$ is the total number of unknown parameters in $\mathscr{W}$ and $\mathscr{B}$. Again, it is important to note that the width of the layer \textit{before} the last linear layer of $f_{\textrm{MLP}}$ approximates the total number of parameters we need to optimize. We set the bottleneck width $r=d+1$ which equals the input dimension of ParameterNet. Hence, the rank of $\mathbf{\Theta}$ is $d+1$. We denote the above dependency of weights and biases simply as $\mathscr{W}(t, \bm{\mu}; \mathbf{\Theta}), \mathscr{B}(t, \bm{\mu}; \mathbf{\Theta})$. By considering \cref{eq:data-fit}, and extending \cref{eq:discrepy_single_snap} from $\mathcal{X}$ to $\Omega$, we arrive at the following minimization formulation,
\begin{equation}
    \label{eq:datafit}
    \min_{\mathbf{\Theta}} \int \mathcal{L} (\mathbf{u}_{\textrm{MLP}}(\mathbf{x}; \mathscr{W}(t, \bm{\mu}; \mathbf{\Theta}), \mathscr{B}(t, \bm{\mu}; \mathbf{\Theta})), 
    \mathbf{u}(\mathbf{x}, t, \bm{\mu})) d\nu(\mathbf{x}, t, \bm{\mu}),
\end{equation}
where $\nu$ now becomes a measure in $\Omega$. A natural choice of $\nu$ to cover the domain of interest is the empirical distribution based on the numerical discretization of $\Omega$ where the data comes from. For example a tensor product of discretizations independently in $\mathcal{X},\mathcal{T}$ and $\mathcal{D}$ leads to $\nu(\mathbf{x}, t,\bm{\mu}) = \sum_{i=1}^{M_{\mathbf{x}}}\sum_{j=1}^{M_t}\sum_{k=1}^{M_{\bm{\mu}}} \delta(\mathbf{x}_i, t_j, \bm{\mu}_k ) / (M_\mathbf{x} M_t M_{\bm{\mu}} )$ where $\delta$ denotes Dirac measure. Here total number of the spatial mesh points is $M_\mathbf{x}$, the number of snapshots in time is $M_t$, and the number of parameters selected is $M_{\bm{\mu}}$. Once a proper $\nu$ is chosen, \cref{eq:datafit} can be solved via gradient-based method, e.g., Adam optimizer \citep{kingma2014adam}. 

It should be noted that \cref{eq:datafit} permits using any kind of proper empirical measure to cover $\Omega$. As shown in \cref{fig1}, this can be especially advantage for problems where an efficient adaptive mesh (e.g., AMR, Overset), moving mesh (e.g., fluid-structure interaction) or simply parameter-dependent mesh (e.g., varying discontinuity with parameters) is adopted. It is a distinct feature that makes NIF different from previous pipelines of dimensionality reduction \citep{xu2019multi,mohan2019compressed} with CAE (i.e., a homogeneous isotropic spatial measure) and SVD (i.e., a tensor product measure structure between $\bm{\mu}$ and $\mathbf{x}$). In this paper, we take $\mathcal{L}$ as the mean square error and rewrite $\nu(\mathbf{x},t,\bm{\nu}) = \sum_{i=1}^{M} \delta(\mathbf{x}_i, t_i, \bm{\nu}_i)/M$ as the most general form\footnote{Taken the example of tensor product measure, the number of total training data points $M$ is the product of resolution on each dimensions, i.e., $M=M_\mathbf{x} M_t M_{\bm{\mu}}$. Therefore, in practice $M$ is typically larger than millions. }. So \cref{eq:datafit} becomes the standard least-squares regression, 
\begin{equation}
\label{eq:datafit_realized}
\min_{\mathbf{\Theta}} \frac{1}{M} \sum_{i=1}^{M} 
\left(\mathbf{u}_{\textrm{MLP}}(\mathbf{x}_i; \mathscr{W}(t_i, \bm{\mu}_i; \mathbf{\Theta}), \mathscr{B}(t_i, \bm{\mu}_i; \mathbf{\Theta}))
- 
\mathbf{u}(\mathbf{x}_i, t_i, \bm{\mu}_i)\right)^2.
\end{equation}

\subsection{Learning representations for multi-scale spatial-temporal data}
\label{sec:ct}

As described in \cref{sec:framework}, ShapeNet is a general MLP with activation function $\sigma$ still to be determined. Given the universal approximator theorem \citep{hornik1989multilayer}, the choice seems to be initially arbitrary. However, there is a significant and often overlooked difference between ``what can MLP approximate'' and ``what can MLP efficiently learn with gradient-based algorithms''. Standard MLPs with common activation functions, such as ReLU, tanh, swish, sigmoid, have been recently observed and proven \citep{tancik2020fourier} to suffer from extreme inefficiency on learning high-frequency functions even with increased network complexity. Indeed, the failure of standard MLPs on high frequency datasets is a well-known phenomenon called \textit{spectral bias} \citep{rahaman2019spectral} or \textit{F-principle} \citep{xu2019frequency}. Recall that in \cref{fig1}, ShapeNet critically relies on an MLP with input $\mathbf{x}$ approximating the ``shape'' of spatial data, which can be problematic in the case of fluid flows with high wavenumber content, e.g., eddies, hairpin vorticies. Note that this issue is also relevant to PINNs \citep{wang2021eigenvector}, which might explain the challenges of using data-free PINNs to solve Naiver-Stokes in fully turbulent regimes.

\begin{figure}[t]
\centering
\includegraphics[width=0.9\textwidth]{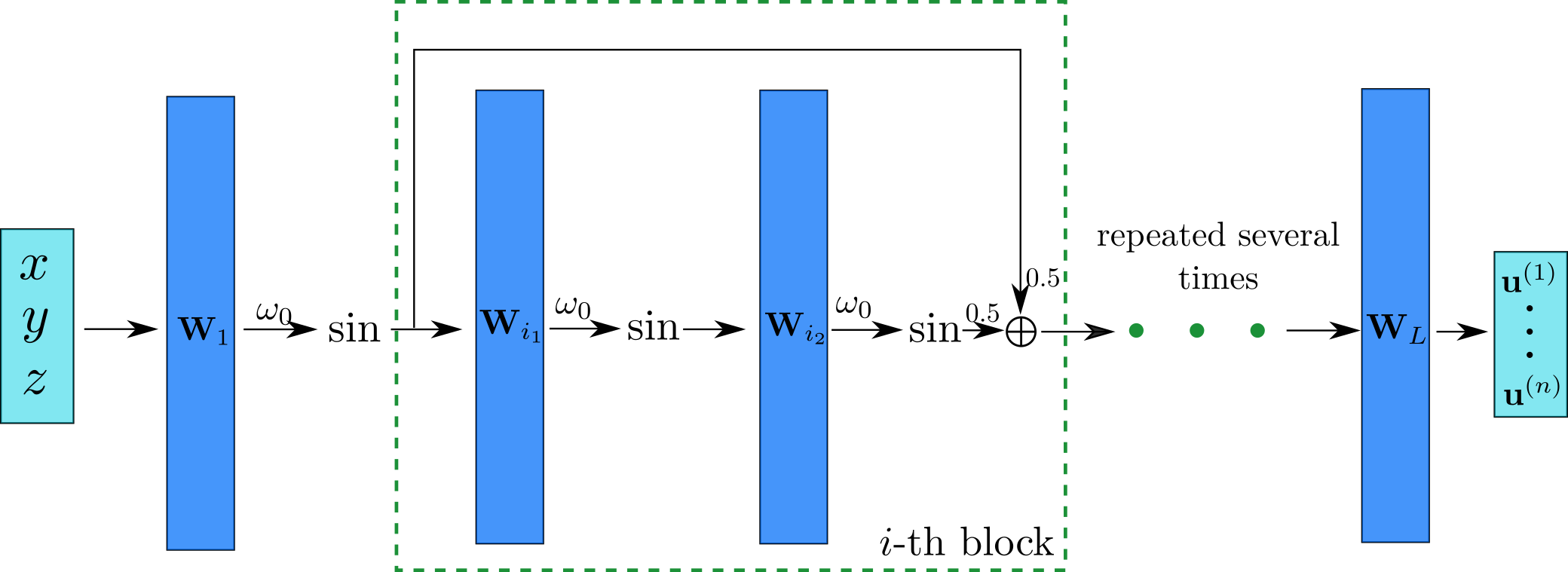}
\caption{ShapeNet for multi-scale spatial data consists of SIREN \citep{sitzmann2020implicit} with ResNet-like structure \citep{lu2021compressive} and $\omega_0$-scaled sine function. Input is spatial coordinate $x,y,z$. Output is a $n$-dimensional vector. Biases are omitted in the figure for clarity. Green dashed line indicates the $i$-th block. Such structure is repeated several times downstream.}
\label{fig:siren_nif_resnet}
\end{figure}

\begin{figure}[t]
\centering
\includegraphics[width=\textwidth]{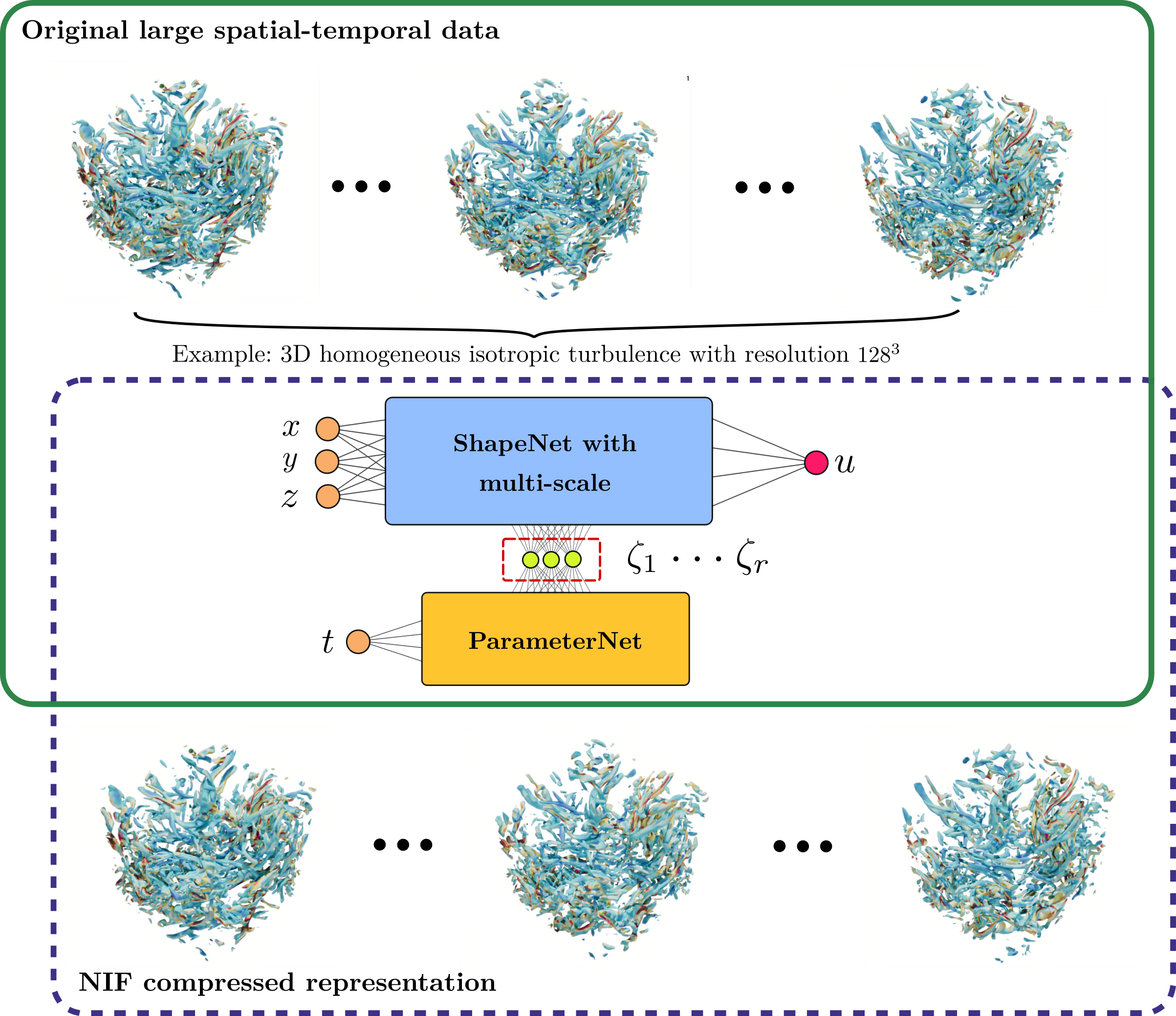}
\caption{Application of NIF with SIREN on mesh-agnostic learning of a latent representation of large-scale spatial-temporal dataset. Iso-contour of Q-criterion  colored by vorticity magnitude of 3D HIT is taken for illustration here and further studied in \cref{sec:app_application_hit}.}
\label{fig:4_compressed_turb}
\end{figure}

Recent advances in computer graphics propose remedies that uses Fourier features \citep{tancik2020fourier} or $\omega_0$-scaled sine activation functions\footnote{Activation function becomes $\sigma(\cdot) = \sin(\omega_0\cdot)$. We use $\omega_0=30$ throughout this work.} in MLP, called {SIREN} \citep{sitzmann2020implicit}, to learn high frequency content. The former Fourier feature approach has been recently introduced in the PINN community by \cite{wang2021eigenvector} with the need to tune the length scale of the Fourier features for each dataset. Here we take the latter approach since we empirically found SIREN is much less sensitive to hyper-parameters of the network compared to the original Fourier feature approach \citep{tancik2020fourier}. Thus, as shown in \cref{fig:siren_nif_resnet}, we design ShapeNet using SIREN with a ResNet-like structure. However, implementing such an $\omega_0$-scaled sine activation function requires a special type of initialization \citep{sitzmann2020implicit} of both the MLP parameters and the uniform normalization of the dataset. Implementation details are documented in \cref{apdx:nif_sine}. 
{\textcolor{black}{With SIREN for the ShapeNet, we can feed time $t$ into ParameterNet to learn a compressed representation from spatial-temporal data in a mesh-agnostic way as illustrated in \cref{fig:4_compressed_turb}.}  Furthermore, instead of feeding time $t$ into ParameterNet, we can build an encoder just using sensor measurements from only a few locations as input to the ParameterNet. Such applications are demonstrated in \cref{sec:app_better_compression,sec:app_application_hit}.}


\subsection{Learning mesh-agnostic linear representations}
\label{sec:linspace}

The {\em proper orthogonal decomposition} (POD) \citep{lumley1967structure} was introduced to the fluid dynamics community by Lumley in 1965 in order to give a mathematical definition of ``large eddy'' by applying a Karhunen-Loeve expansion \citep{loeve1955probability} to turbulent velocity fields. Formally, the first $r$ POD modes are determined by the first $r$ eigenfunctions of an integral operator with the cross-correlation kernel of the turbulent velocity field \citep{george1988insight}. In practice, the SVD is often used to obtain a discrete realization of POD modes \citep{brunton2019data}. Such POD modes not only find an $r$-dimensional linear subspace that optimally minimizes the $L^2$ projection error (as shown in \cref{eq:pod}) but also provides a sequence of ordered basis weighted by their singular values \citep{djouadi2008optimality}.
\begin{equation}
\label{eq:pod}
\min_{\bm{\psi}_1,\ldots,\bm{\psi}_r} \iint \sum_{l=1}^{n} \left( \mathbf{u}^{(l)}(\mathbf{x},t) - \sum_{i=1}^{r} \alpha_i(t) \bm{\psi}^{(l)}_{i}(\mathbf{x}) \right)^2 d\mathbf{x}dt,
\end{equation}
subject to the constraints $\alpha_i(t) = \int \sum_{l=1}^{n} \mathbf{u}^{(l)}(\mathbf{x},t) \bm{\psi}^{(l)}_{i}(\mathbf{x})  d\mathbf{x}$, $\int \sum_{l=1}^{n} \bm{\psi}^{(l)}_{i}(\mathbf{x}) \bm{\psi}^{(l)}_{j}(\mathbf{x})d\mathbf{x} =\delta_{ij}$, for $i=1,\ldots,r$, where the superscript $l$ denotes $l$-th component and $\delta_{ij}$ is the Kronecker delta.
POD via SVD relies on a fixed mesh in order to provide a closed-form discrete approximation of the POD modes. If the mesh changes with time and/or parameters, which is the cases for many problems of interest \citep{teyssier2002cosmological,bryan2014enzo,vay2004application}, then the SVD-based approaches are ill-suited for many downstream tasks such as modal analysis of fluid flows \citep{taira2017modal} and reduced-order modeling \citep{benner2015survey, loiseau2021pod, noack2003hierarchy}. 

\begin{figure}[t]
\vspace{-.2in}
\centering
\includegraphics[width=0.8\textwidth]{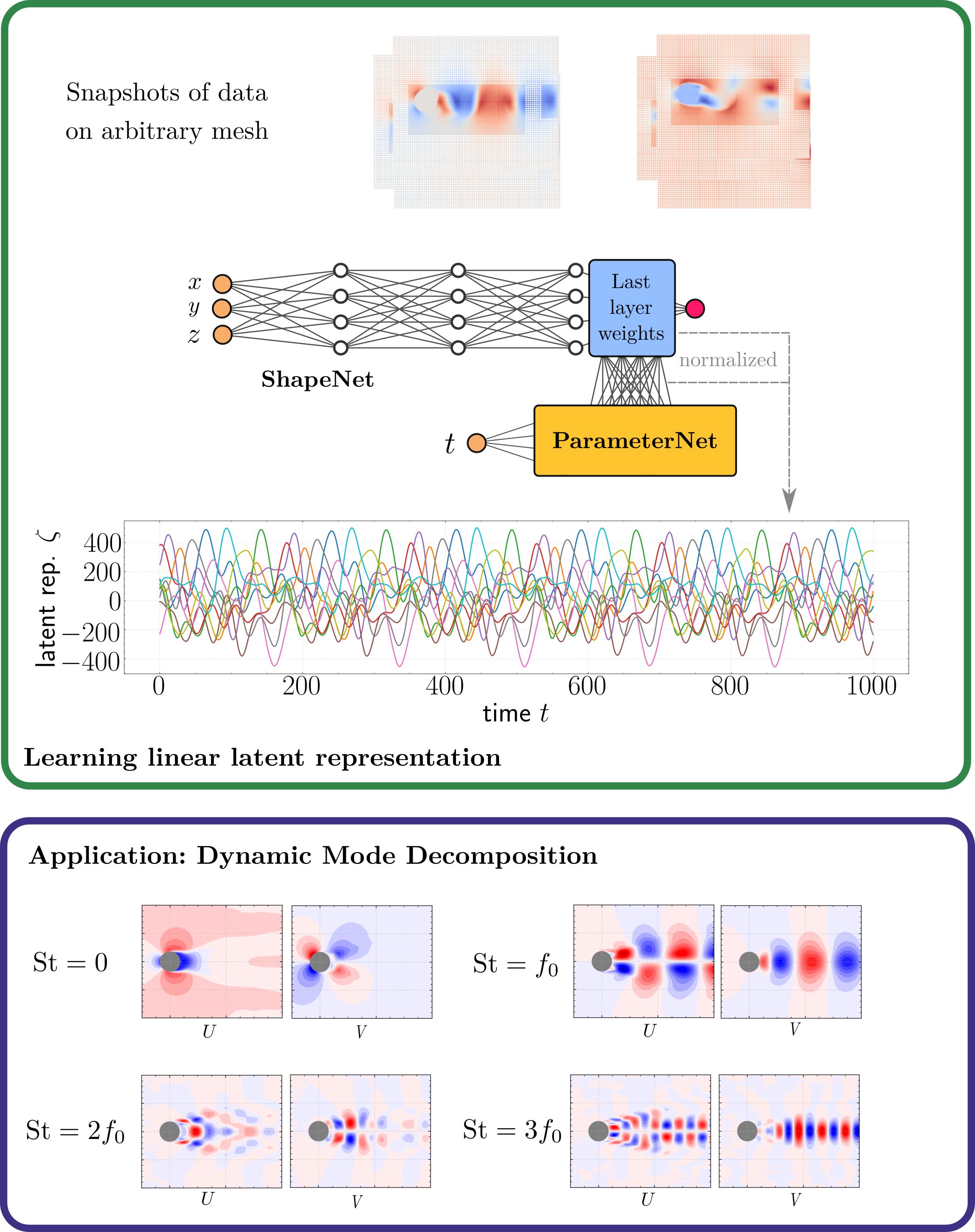}
\vspace{-.1in}
\caption{Application of NIF on mesh-agnostic learning of linear subspace. Here a flow over cylinder at $Re \approx 123$ is used for illustration. Once latent subspace is learned from spatio-temporal dataset with AMR. A standard DMD \citep{brunton2019data} is performed on the latent representation. $f_0$ denotes the fundamental frequency. When postprocessing the DMD mode shape, we choose a Cartesian grid with 500 uniform sampling points in each direction.}
\label{fig:linear_subspace_cyd}
\vspace{-.1in}
\end{figure}

Since multi-scale features often appear in spatio-temporal data, we employ NIF with SIREN in \cref{sec:ct} for the applications considered in the rest of this paper.
As shown in \cref{fig:linear_subspace_cyd}, we first provide a framework based on NIF to \textit{directly} approximate an optimal $r$-dimensional linear space of classical POD theory \citep{djouadi2008optimality}. The key observation is that we can use ParameterNet with input $t$ to parameterize only the last layer weights and biases of ShapeNet while the rest of weights and biases of ShapeNet are determined by optimization. As shown in \cref{eq:nif_linss}, we arrive at an interpretable approximation of the original spatio-temporal field $\mathbf{u}(\mathbf{x},t)$ as a sum of $r$ products of spatial functions $\bm{\phi}_1,\ldots,\bm{\phi}_r$ and temporal modes $a_1,\ldots,a_r(t)$ parameterized by MLP,
\begin{equation}
\label{eq:nif_linss}
\min_{\mathbf{\Theta}, \{\mathbf{W}_i, \mathbf{b}_i\}_{i=1}^{L-1}} \iint \sum_{l=1}^{n} \left( \mathbf{u}^{(l)}(\mathbf{x},t) - \sum_{i=1}^{r} a_{\textrm{MLP},i}(t; \mathbf{\Theta}) \bm{\phi}^{(l)}_{\textrm{MLP},i}(\mathbf{x}; \{\mathbf{W}_j, \mathbf{b}_j\}_{j=1}^{L-1} ) \right)^2 d \mathbf{x} dt.
\end{equation}
The notations in \cref{eq:nif_linss} is slightly different from \cref{eq:pod} in order to highlight that NIF only approximates the $r$-dimensional linear subspace rather than obtaining a set of ordered orthonormal spatial functions. \textcolor{black}{Note that one needs to take the cell area into account when implementing $\cref{eq:nif_linss}$. To remove the effects of the slightly varying magnitude of the spatial basis learned with a neural network, we consider a normalizing spatial basis, such that $\tilde{\phi}_{\textrm{MLP},i}^{(l)} = \phi_{\textrm{MLP},i}^{(l)}/c_i$ where $c_i \triangleq \sqrt{\int \sum_{l=1}^{n}(\phi_{\textrm{MLP},i}^{(l)}(\mathbf{x}))^2 d\mathbf{x}}$ and $ \zeta_i(t) \triangleq c_i a_{\textrm{MLP},i}(t)$.
Since $\zeta(t) \in \mathbb{R}^{r}$} is the corresponding mesh-agnostic $r$-dimensional linear representation, one can effortlessly apply any existing SVD-based frameworks (e.g., SVD-DMD \citep{schmid2010dynamic}) on datasets with arbitrary varying meshes.

\subsection{Mesh-agnostic data-driven non-linear sparse reconstruction}
\label{sec:ss}

The goal of data-driven sparse reconstruction is to use limited sensor measurements to infer the entire high-dimensional system state, given {\em a priori} information of the low-dimensional manifold where system evolves. It has been widely applied in projection-based reduced order modeling, especially  for large-scale nonlinear systems (also known as ``hyper-reduction'') where computing expensive nonlinear terms can be avoided. Currently POD-QDEIM~\citep{drmac2016new} is one of the most popular methods, which shows improved performance over classical compressive sensing techniques~\citep{manohar2018data}. The idea of POD-DEIM~\citep{chaturantabut2010nonlinear} is to use least-square estimators for the latent representation by only measuring a few locations (e.g., sensors). \citet{chaturantabut2010nonlinear} derived an error upper-bound of DEIM that indicates two contributions: 1) a spectral norm related to sensor selection and 2) projection error of the linear subspace. The idea of POD-QDEIM is to minimize the former contribution of the spectral norm with QR pivoting. 
Without loss of generality, given a scalar field $u$ on $M_{\mathbf{x}}$ mesh points and $M_t$ snapshots, the data matrix is 
\begin{equation}
\mathbf{U}= 
\begin{bmatrix}
u(\mathbf{x}_1, t_1) & \ldots & u(\mathbf{x}_1, t_{M_t}) \\ 
\vdots & \vdots & \vdots \\ 
u(\mathbf{x}_{M_{\mathbf{x}}}, t_1) & \ldots & u(\mathbf{x}_{M_{\mathbf{x}}}, t_{M_t}) 
\end{bmatrix} \in \mathbb{R}^{M_{\mathbf{x}} \times M_t}.
\end{equation}
The corresponding reduced rank-$r$ SVD is $\mathbf{U} \approx \mathbf{\Psi}_r \mathbf{\Sigma}_r \mathbf{V}_r^\top$. Near-optimal sensor placement can be achieved via QR factorization with column pivoting of the transpose of spatial basis,
\begin{equation}
\label{eq:qdeim}
\mathbf{\Psi}_r^\top \mathbf{C}^\top = \mathbf{Q} \mathbf{R}.
\end{equation}
Note that the top $p$ rows of $\mathbf{C}=\begin{bmatrix}
\mathbf{e}_{\gamma_1} & \ldots & \mathbf{e}_{\gamma_p}
\end{bmatrix}^\top \in \mathbb{R}^{p \times M_{\mathbf{x}}}$ give a sparse measurement matrix, where $\mathbf{e}_i$ are the canonical basis vectors with a unit entry at index $i$ and zero elsewhere. $\gamma_i$ corresponds to the index of $i$-th best sensor location.  One can recover the full field by least-square estimation of the latent representation from $u$ at those sensors. 

\begin{figure}[t]
\centering
\includegraphics[width=0.8\textwidth]{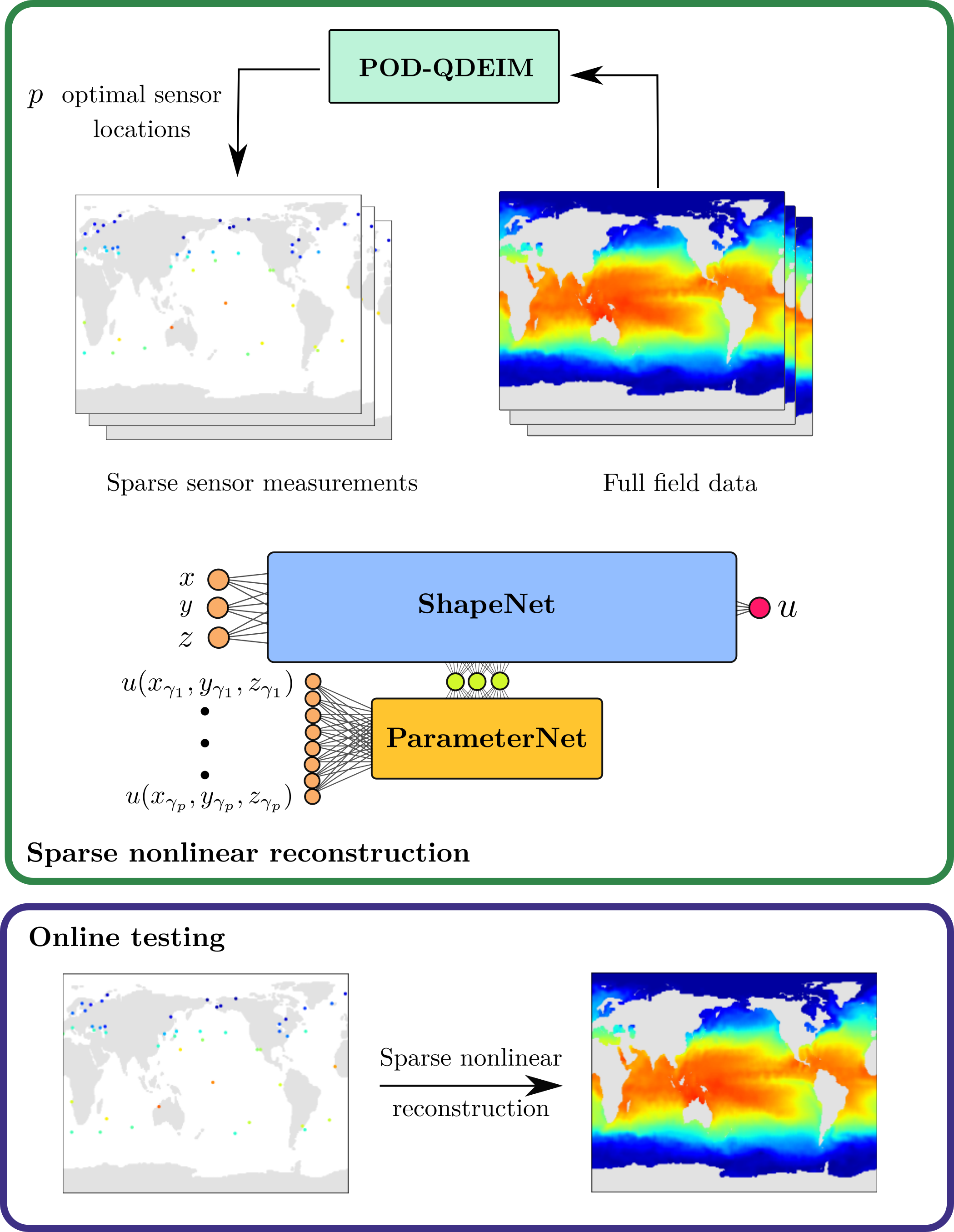}
\caption{Application of NIF on mesh-agnostic data-driven sparse reconstruction.}
\label{fig:nif_sparse_sensing}
\end{figure}

Given the success of POD-QDEIM, we can turn our attention to its error contribution, i.e., the projection error of the linear subspace, by replacing POD with NIF. We first use the optimized sensor location determined by POD-QDEIM. Then, as shown in \cref{fig:nif_sparse_sensing}, given $p$ sensor measurements $u(\mathbf{x}_{\gamma_{1}}), \ldots, u(\mathbf{x}_{\gamma_{p}})$ as input for ParameterNet, and the ground true field $u$ at $\mathbf{x}$, we end up with a standard supervised learning problem. Once the model is trained, it can be used to predict the full spatial field $u$ at any location for a given measurement from $p$ sensors. 

\section{Applications}
\label{sec:applications}

The code and data for the following applications is available at \url{https://github.com/pswpswpsw/paper-nif}. The Python package for NIF is available at \url{https://github.com/pswpswpsw/nif}.

\subsection{Learning parametric solutions of Kuramoto–Sivashinsky equation}
\label{sec:app_data-fit-1d-ks}

We apply the data-fitting parametric surrogate modeling framework in \cref{eq:datafit_realized} on the 1D Kuramoto–Sivashinsky equation with periodic boundary condition, 
\begin{equation}
\label{eq:ks}
u_t + uu_x + u_{xx} + \mu u_{xxxx} = 0, \quad u(0,t)=u(2\pi, t), 
\end{equation}
with varying parameter $\mu$ as shown in \cref{fig-r1}. 
For the K-S equation, we fix the initial condition as $\sin(x)$ and vary $\mu$ from 0.2 to 0.28 which is not chaotic\footnote{Although most often the K-S equation is simulated on chaotic regime, it is also famous for rich bifurcation phenomenon as its parameters change \citep{papageorgiou1991route}.}. The training data consists of 20 points in the parameter $\mu$ space (i.e., 20 simulations with distinct $\mu$). The testing data consists of 59 simulations with a finer sampling of $\mu$. As shown in \cref{fig-r1}, the system response when one varies $\mu$ from 0.2 to 0.28 is relatively smooth without any chaos. This makes it a well-posed problem for regression. The training data is preprocessed with standard normalization. Details are given in \cref{apdx:ks}.

For NIF, we take 4 layers with units for ParameterNet as 2-30-30-2-6553 and 5 layers with units 1-56-56-56-1 with ResNet-like skip connection for ShapeNet. We empirically found such ResNet-like skip connections can help accelerate the convergence. Note that 6553 corresponds to the total number of weights and biases in the aforementioned ShapeNet. The swish activation function \citep{ramachandran2017searching} is adopted.
As a comparison, we use a standard MLP with 5 layers with units 3-100-100-100-1, which is around the same number of model parameters with $x,t,\mu$ as input and output $u$. This can be viewed as a straightforward idea using PINNs~\citep{raissi2020hidden} as the regression without minimizing the PDE loss. Note that the same skip connections are employed as well. The model parameters are initialized with a truncated normal with standard deviation of 0.1 for both cases\footnote{We empirically find Such ``small weights initialization'' is an easy way to help the convergence of NIF. Systematic work in \citep{chang2019principled} on initialization of hypernetwork might further improve the result.}. For the standard MLP case, two extra cases with tanh and ReLU activations are considered. We implemented both models in Tensorflow \citep{tensorflow}. Note that we heavily rely on \texttt{einsum} in implementing NIF. We adopt the Adam optimizer \citep{kingma2014adam} with a learning rate of 1e-3, batch size of 1024 and 40000 epochs. To 
take training randomness into account and remove outliers, we take the average of 4 well converged trials for both network structures. As shown in \cref{tab:datafit-1d-pde-error}, NIF with Swish activations achieved better performance on both training and testing data than three MLP counterparts. 

\begin{table}[bthp]
\centering
\scriptsize
\caption{Comparison between standard MLP and NIF in \cref{sec:datafit} for surrogate modeling of 1D parametric PDE. RMSE below is averaged over all parameter $\mu$.}
\label{tab:datafit-1d-pde-error}
\vspace{1em}
\begin{tabular}{ccc}
\textbf{Model}   & \textbf{Train RMSE} & \textbf{Test RMSE} \\ 
\midrule
NIF (Swish) & $\textbf{1.9} \bm{\times 10^{-2}}$ & \textbf{0.64}  \\
MLP (Swish) & $2.2 \times 10^{-2}$ & $1.10$ \\
NIF (tanh) & $3.9 \times 10^{-2}$ & 0.99  \\
MLP (tanh)  & $3.7 \times 10^{-2}$ & $1.17$  \\
MLP (ReLU)  & $7.0 \times 10^{-2}$ & $1.27$ \\
\bottomrule
\vspace{1em}
\end{tabular}
\end{table}

\begin{figure}[bthp]
\centering
\includegraphics[width=\textwidth]{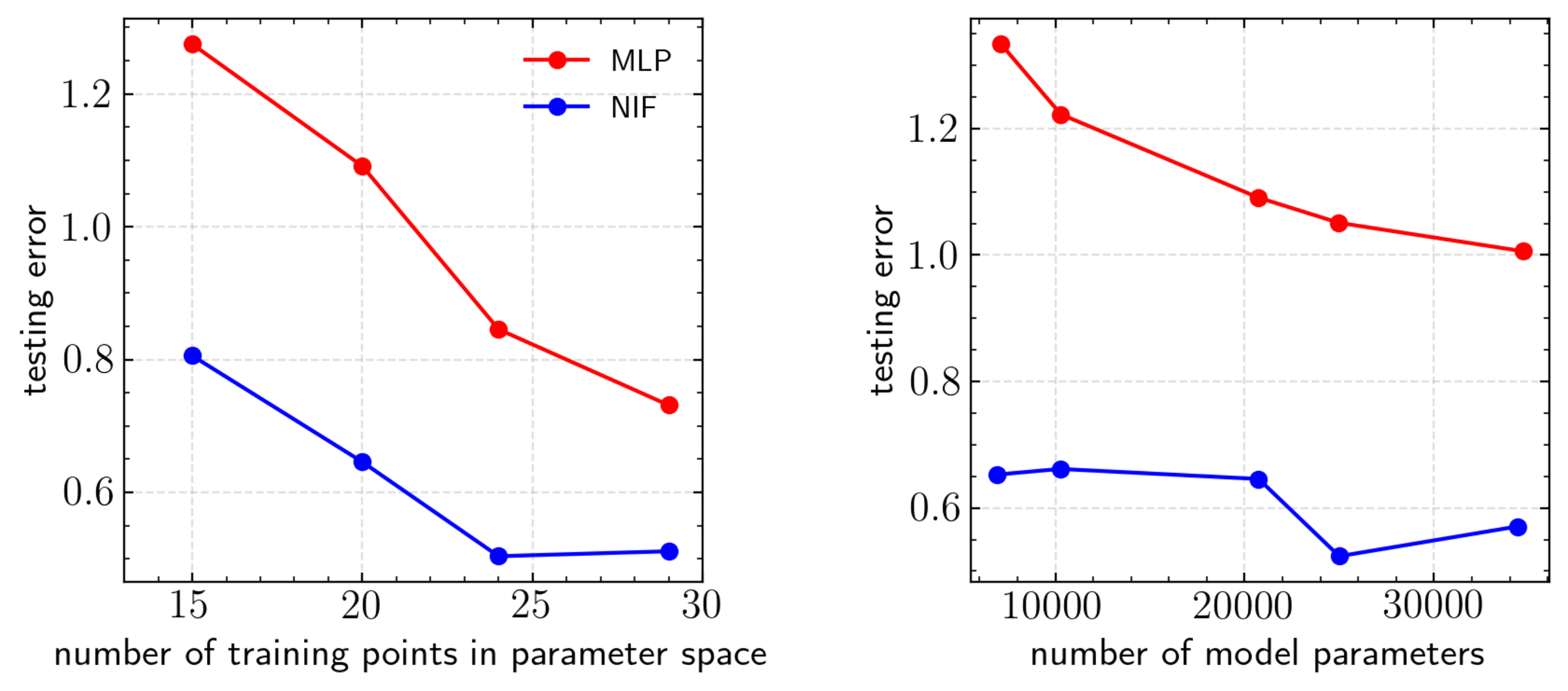}
\caption{Trend of testing error against (Left) varying the size of training data; (Right): varying the number of model parameters.}
\label{fig:ks_0_change_data_mp}
\end{figure}

For simplicity, we fix the Swish activation function in the following. We first vary the size of the training data and retrain the models to evaluate data efficiency. We change the number of sampling points in parameter space for training data from the previous 20 to 15, 24, 29. As shown in \cref{fig:ks_0_change_data_mp}, NIF with Swish activation performs consistently better than MLP with Swish activation. To achieve the same level of testing error, NIF requires approximately half of the training data. Finally, we vary the number of model parameters from 7000 to 34000 while fixing the number of points in parameter space to 20.  We then retrain the models to evaluate model expressiveness. As displayed in \cref{fig:ks_0_change_data_mp}, given the same number of parameters, NIF lowers the testing error by half compared to its MLP counterpart. Details of the comparisons are given in \cref{apdx:datafit}.

\subsection{Nonlinear dimensionality reduction of Rayleigh-Taylor instability from adaptive mesh}
\label{sec:app_better_compression}


To highlight the advantage of NIF on multi-scale problems, here we compare the performance of SVD, CAE and a NIF-based autoencoder on reducing the dimensionality of the density field from the classical 2D Rayleigh-Taylor (R-T) instability. In this R-T problem, the interface between the high density fluid above and the low density fluid below is initially perturbed with a single mode profile of density. The CFD simulation is performed with CASTRO~\citep{almgren2010castro}, which is a compressible hydrodynamic code with AMR. 
Since the mesh changes with time, CAE or SVD can only be applied after projecting the data from the adaptive mesh onto a static fine mesh. Such preprocessing can introduce the so-called projection error and can be computationally challenging, especially in 3D. In the following, we take the static fine mesh as 128$\times$256 for SVD and CAE. In contrast, NIF directly takes the pointwise raw data on the adaptive mesh for training. 

The goal here is to encode the flow state onto an $r$-dimensional latent space, from which one can faithfully reconstruct the flow state. Note that $r$ is the dimension of the latent subspace. Since we only take data from a single initial condition and the collection of data ends before the symmetry breaks, the minimal latent space dimension $r$ that preserves the information in this case is one. We sample the flowfield uniformly in time and split such single trajectories into 84 training and 28 testing snapshots in a way that the testing snapshots fall in between the training snapshots. Details of data preparation are provided in \cref{apdx:rt}. 

Note that the structure of NIF in \cref{fig1} is only for a \textit{decoder} rather than an \textit{encoder}. Hence, unlike SVD and CAE, we still need to choose an encoder that feeds information to the network in order to let it discern one snapshot from the other. One can choose either a convolution layer with coarse data on Cartesian meshes just like CAE, or \textcolor{black}{a neural network with} a cluster of point-wise measurements at certain locations. Here we choose the latter: we consider 32 uniformly distributed sensors along the vertical axis in the middle of the flowfield. 

For NIF, we take two ResNet blocks with 64 units in \cref{fig:siren_nif_resnet} followed by a linear bottleneck layer of $r$ units as ParameterNet.  ShapeNet contains 2 ResNet-like blocks in \cref{fig:siren_nif_resnet} with 128 units. ParameterNet takes the 32 sensor measurements as input and outputs 66,561 parameters as the weights and biases of the ShapeNet. While ShapeNet takes pointwise $(x,y)$ coordinates as input and outputs the prediction of density $u$ at $(x,y)$.
The output dimension of ParameterNet is 66,561 as the total number of weights and biases of the ShapeNet. \textcolor{black}{To enforce a smooth latent representation, we use Jacobian and approximated Hessian regularization~\citep{rifai2011higher} together with an initialization of small weights for the encoder, which we find empirically to be helpful.}

For CAE, We choose a typical deep convolutional architecture used for fluid flows \citep{wiewel2019latent} with detailed setup in \cref{apdx:cae-details}. Gradient-based optimization is performed with an Adam optimizer. The learning rate is 2e-5 for NIF and 1e-3 for CAE with a batch size of 3150 for NIF and 4 for CAE\footnote{Note that pointwise data is fed to NIF while image snapshot data projected on a uniform Cartesian mesh is fed to CAE.}. The total learning epoch is 10,000 for CAE and 800 for NIF. 

\begin{figure}[htbp]
\centering
\includegraphics[width=\textwidth]{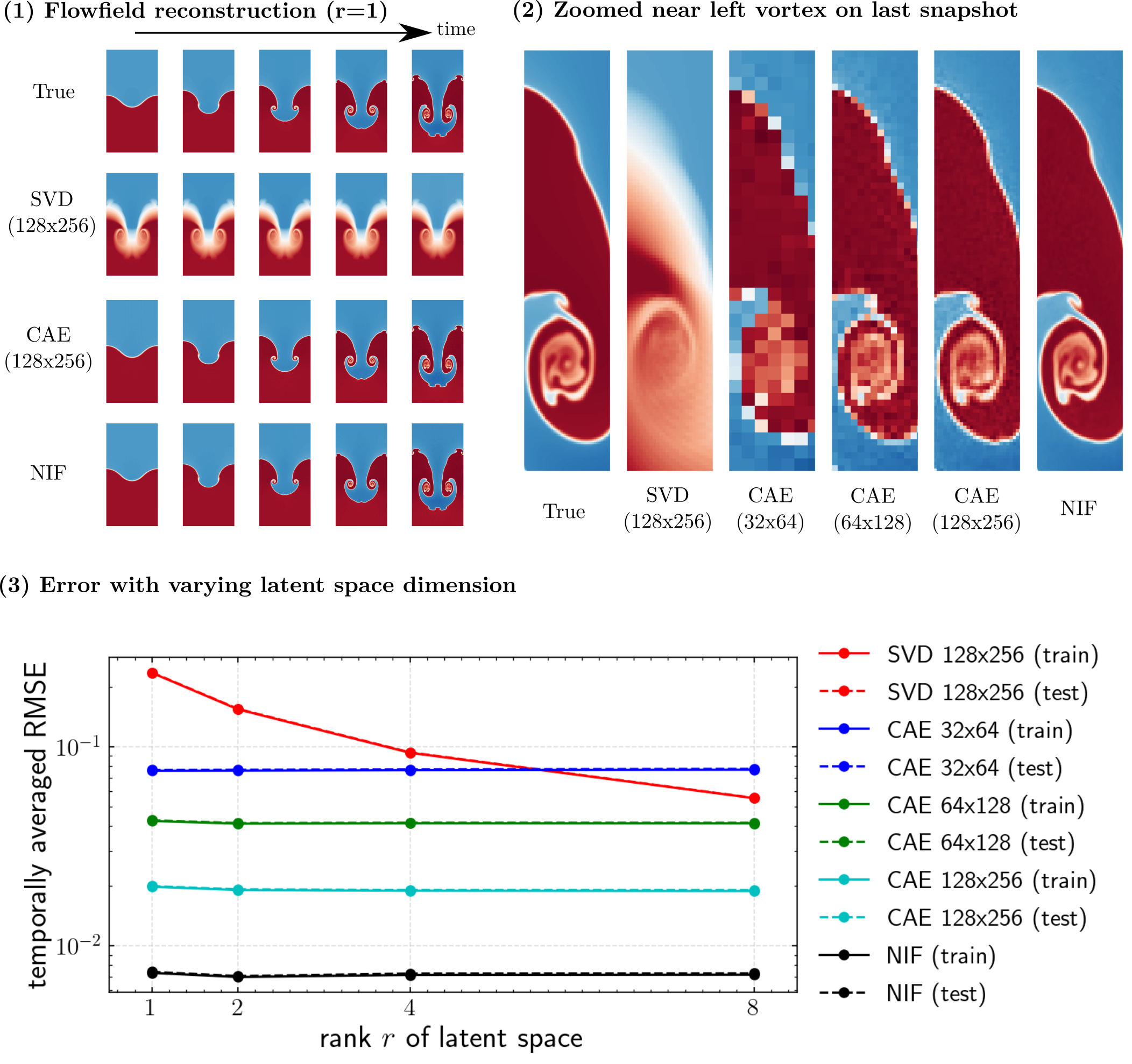}
\caption{Comparison of SVD, CAE with 32x64 (low), 64x128 (middle) and 128x256 (high) resolution for dimensionality reduction on testing data of 2D Rayleigh-Taylor instability on an adaptive mesh. (1) evolution of flowfield reconstruction with $r=1$ at five selected time in the testing phase. (2) Zoomed near the left vortex on the last snapshot. (3) Trend of temporally averaged RMSE with varying rank $r$ for all models.}
\label{fig:rt-compare-autoencoder}
\end{figure}

In order to make quantitative comparisons, we project all of the predictions on testing data together with ground true data onto a very fine mesh with resolution of 256$\times$512 by nearest-neighbor (for CAE and SVD) or direct sampling (for NIF).
As shown in \cref{fig:rt-compare-autoencoder} where a rank 1 reduction is performed, the prediction of NIF is better than SVD and CAE with varying static mesh resolution from $32\times 64$,  $64\times 128$ and  $128\times 256$. When $r$ increases from $1$ to $8$, it is expected that the errors from non-linear models do not change much due to the nature of single realization while that from the linear method decreases. On average, predictions of the CAE with three increasing resolutions lead to 10, 5, 1.7 times more error than that of NIF model. Further error analysis in \cref{apdx:compare_time_evolution_err} shows that the NIF-based autoencoder outperforms SVD mainly due to the lack of nonlinear expressiveness of SVD and it is better than CAE because of the excess projection error from the uniform Cartesian grid.

\subsection{Learning spatially compressed representation for 3D homogeneous isotropic turbulence}
\label{sec:app_application_hit}

Next, we apply NIF in \cref{sec:ct} to learn a spatially compressed representation of 3D multi-scale spatio-temporal field. Our goal is to find a vector-valued continuously differentiable function $\mathbf{f}_{\textrm{MLP}}(\mathbf{x}; \mathbf{\Theta}(t))$ which ``fits'' the original spatial-temporal data. $\mathbf{\Theta}(t)$ is linearly determined by the time-dependent reduced coordinates $\zeta_1,\ldots,\zeta_r$. Note that $r$ is several order of magnitude smaller than the number of mesh points. If it is achieved, one can efficiently send a snapshot of turbulence data at any time $t$ by just transferring a $r$-dimensional vector $\mathbf{\Theta}(t)$ to the receiver. While the receiver just needs a ``skeleton'' ShapeNet and a single linear decoder layer (i.e., the last layer of ParameterNet) at local device in order to decode $\mathbf{\Theta}(t)$ from the sender into a continuous differentiable spatial field. It is important to note that the last layer of ParameterNet is a very wide linear layer, with the width on the order of the total number of weights and biases of ShapeNet.

As an illustrative example, we use a temporally evolving (20 snapshots) spatial ($128^3$) velocity field of homogeneous isotropic turbulence (HIT) from Johns Hopkins University. Details of data preparation are given in \cref{apdx:3dhit}. 
It should be highlighted that distinct from CAE or SVD, the model complexity of NIF is not directly related to the resolution that the data is stored but rather the \textit{intrinsic} spatial complexity of the data itself. In this example, as for network architecture, ShapeNet has 4 ResNet-like blocks as hidden layers with width as 200 while ParameterNet has 1 ResNet-like block with width as 50 followed by a linear layer with $r=3$ width and a linear layer that maps the $r$-dimensional vector to all the weights and biases of ShapeNet. The total number of trainable parameters is 1,297,365, which is only inside ParameterNet. For training, we use an Adam optimizer with a learning rate of 1e-5 and batch size of 1600. 

\begin{figure}[htbp]
\centering
\includegraphics[width=0.9\textwidth]{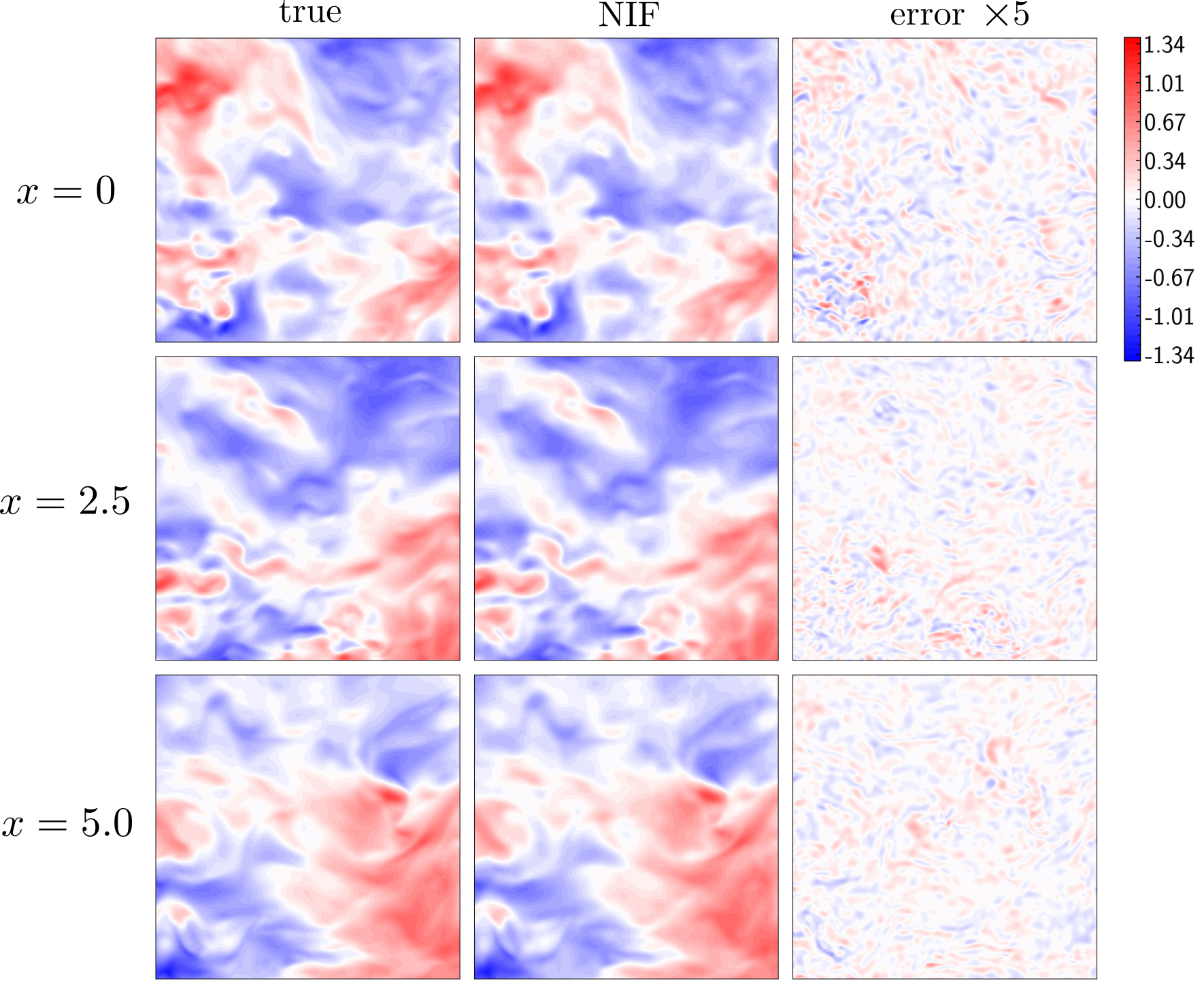}
\caption{Evaluation of compressed reconstruction from NIF on the $U$ velocity field of the three dimensional homogeneous isotropic turbulence at three difference slices ($x_{\textrm{max}}=9.9$) and $t=0$ from JHU dataset~\citep{li2008public}. Left: ground true sliced $U$ field. Middle: reconstruction from NIF. Right: amplified error between ground true and reconstruction from NIF.}
\label{fig:hit_compare_U}
\end{figure}

\begin{figure}[htbp]
\centering
\includegraphics[width=0.9\textwidth]{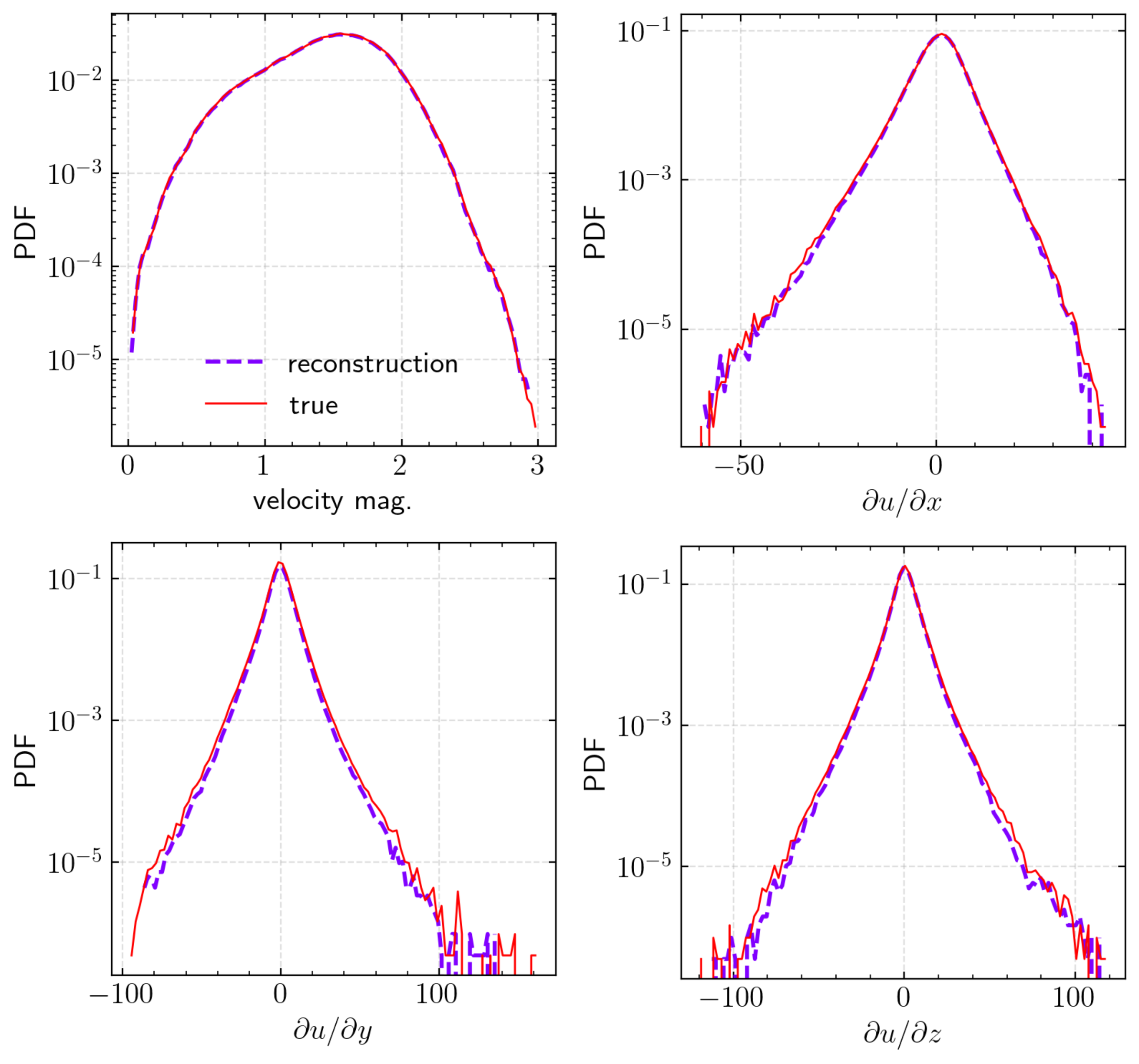}
\caption{Evaluation of compressed reconstruction from NIF on the PDF of velocity magnitude, $\partial u/\partial x$, $\partial u/\partial y$, $\partial u/\partial z$ of the 3D HIT from JHU dataset~\citep{li2008public} at the first time instance.  
}
\label{fig:hit_compare_PDF}
\end{figure}

\begin{figure}[htbp]
\centering
\includegraphics[width=\textwidth]{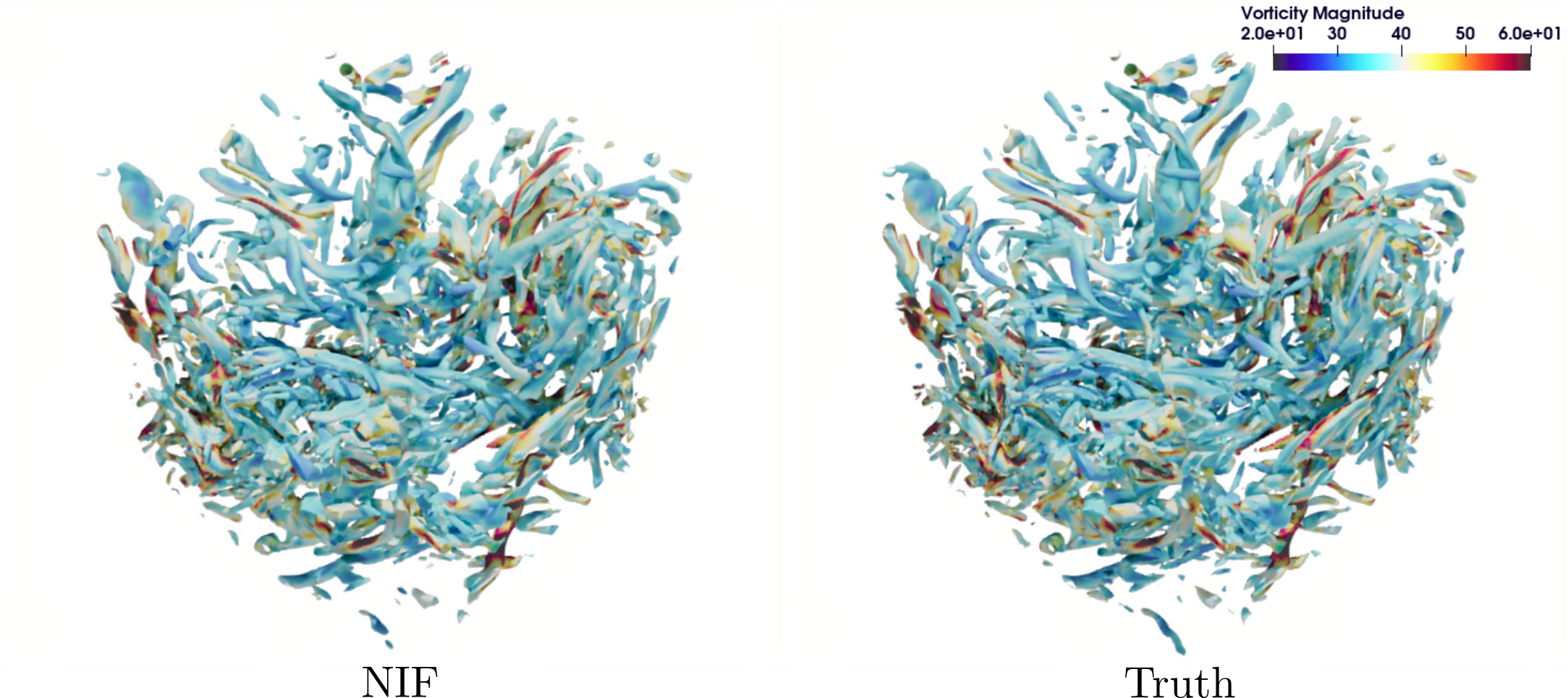}
\caption{Comparison of compressed reconstruction of 3D HIT (left) and ground true at first snapshot (right). Flowfield is visualized with a contour of Q-criterion colored with vorticity magnitude.}
\label{fig:hit_q}
\end{figure}

First, we test our model by comparing the first component of the ground true velocity field versus the reconstruction from NIF. As displayed in \cref{fig:hit_compare_U}, NIF reconstructs the ground true velocity even with small-scale structures very well.
Since most modern studies on turbulence are descriptive from a statistical viewpoint, it is important to verify that the PDF is well preserved after compression. As shown in \cref{fig:hit_compare_PDF}, the PDFs of various quantities are approximately well preserved.
For a more stringent test, we verify the model performance by visually comparing the iso-contour of Q-criterion and vorticity magnitude. As displayed in \cref{fig:hit_q}, most of the high order quantity is well preserved by the model with only small visual difference. 
Lastly, it is important to highlight that the original dataset require a storage of an array with size $128^3\times 20 = 41,943,040\approx$ while the total number of parameters need to be trained is $1,297,365$ which is $3\%$ of the former. Further, such compression ratio can be improved with more recent neural network pruning techniques~\citep{han2015deep} and more inputs to ParameterNet. We leave this exciting topic for future study.

\subsection{Efficient spatial query of 3D homogeneous isotropic dataset} 
\label{sec:app_efficient_spatial_query}

When we are querying large scientific datasets,  typically from PDE solvers, we perform many more queries in space than queries in time or parameter space. For example, spatial statistics in the homogeneous direction are often collected in the analysis of turbulent physics. Visualization of vortex indicators requires intensive spatial query, e.g., spatial differentiation. The primary reason is that the spatial degree of freedom is typically much larger than either the temporal or parametric degrees of freedom. 

Since the structure of NIF isolates the spatial complexity independently from any other factors, we can efficiently get the spatial field data without unnecessary \textit{repeated} computations related to other factors such as time or system parameters. On the contrary, it could be a waste of resources if one uses a single feedforward neural network with multiple SIREN layers~\citep{sitzmann2020implicit} that takes \emph{all} information as input with the output still being the field of interests (here we refer it simply as ``SIREN''). It is because such a single network will need to mix and learn all of the complexities, e.g., multi-scale, chaos and bifurcations. While the number of spatial query is typically on the order of 10,000 in 2D and 1 million in 3D, one has to repeatedly perform temporal and/or parametric query for different spatial points if they adopt a single network with \emph{all} information as input for spatial query intensive tasks, e.g., learning representation of video~\citep{sitzmann2020implicit} or temporally evolving volumetric field~\citep{lu2021compressive}. Therefore, this can lead to a potentially larger network with longer inference time under the same level of accuracy.

\begin{figure}[htbp]
\centering
\includegraphics[width=1\textwidth]{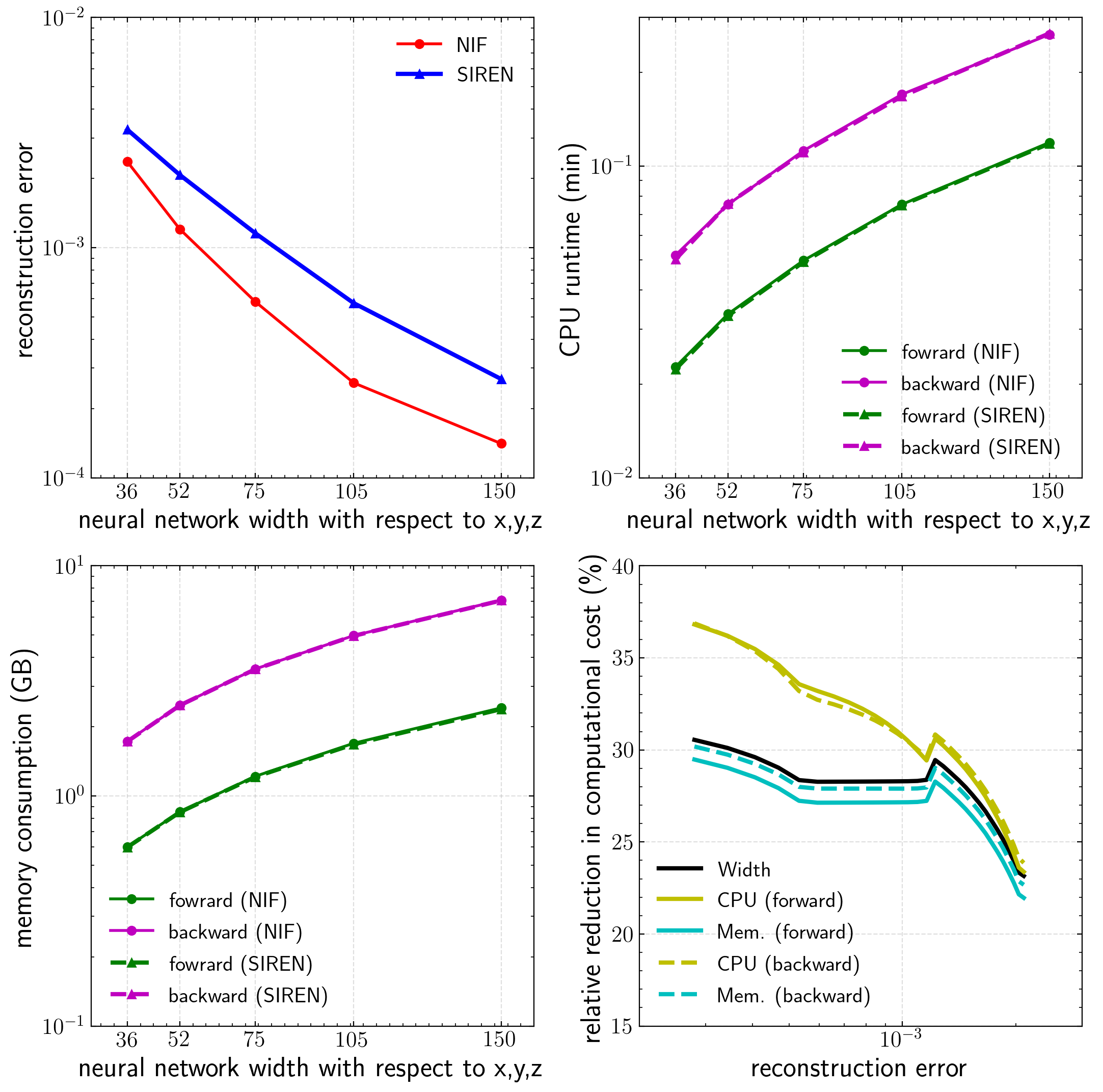}
\caption{Efficiency comparison between NIF and SIREN in term of CPU runtime and memory consumption when performing spatial query test on half of the first snapshot of the $128^3$ homogeneous isotropic turbulence. Top left: variation of reconstruction error with network size; Top right: variation of CPU runtime with network size; Bottom left: variation of memory consumption with network size. Bottom right: relative reduction in computational cost of NIF compared with that of SIREN. Note that the reconstruction error is computed based on the mean-squared error of last batch for every epoch with data shuffling.}
\label{fig:efficient_query}
\end{figure}

Since once the output of ParameterNet is given, the spatial inference can be performed with only run inference on the ShapeNet.  We use the same HIT data with $128^3$ resolution (see \cref{sec:app_application_hit}). Our model setup is the same as before except that the width of the ShapeNet and that of the SIREN change from 36, 52, 75, 105 to 150 and the width of the ParameterNet increases from 3 to 10. As shown in the top left of \cref{fig:efficient_query}, NIF uses a \textit{smaller} network for spatial evaluation compared to SIREN counterpart under the same level of reconstruction error. This is because SIREN takes $t,x,y,z$ as input so capacity of the network is also spent on learning temporal variation. 

To further compare the performance, we measure CPU runtime and memory consumption of the spatial query part in the code with the  standard Python package \texttt{time} and \texttt{Fil-memory-profiler}~\citep{filprofiler}. We denote the spatial query of three velocity components as a \emph{forward} computation while $\partial u/\partial x$ is denoted as a \emph{backward} computation. To make it a fair comparison, for NIF we take the inference time on ParameterNet (although it contributes less than 1\% to the total computational time here) into consideration as well. Note that in the top right and bottom left subfigures of \cref{fig:efficient_query}, there is not much difference in terms of the scaling between NIF and SIREN, which is simply the similarity between the ShapeNet and the SIREN except the first layer of SIREN takes $t,x,y,z$ as inputs while that of NIF takes $x,y,z$. We also note that the backward computation requires nearly twice the computational cost as that of forward computation. However, with the same accuracy NIF requires a smaller network than SIREN. Hence, given the same reconstruction error, the width required for NIF and SIREN can be determined. The bottom right of \cref{fig:efficient_query} indicates that NIF leads to 27\% less neural network width, 30\% less CPU time in forward and backward passes and 26\%/27\% less memory consumption in forward/backward computations for the task of querying homogeneous isotropic turbulence data. 

\begin{figure}[htbp]
\centering
\includegraphics[width=1\textwidth]{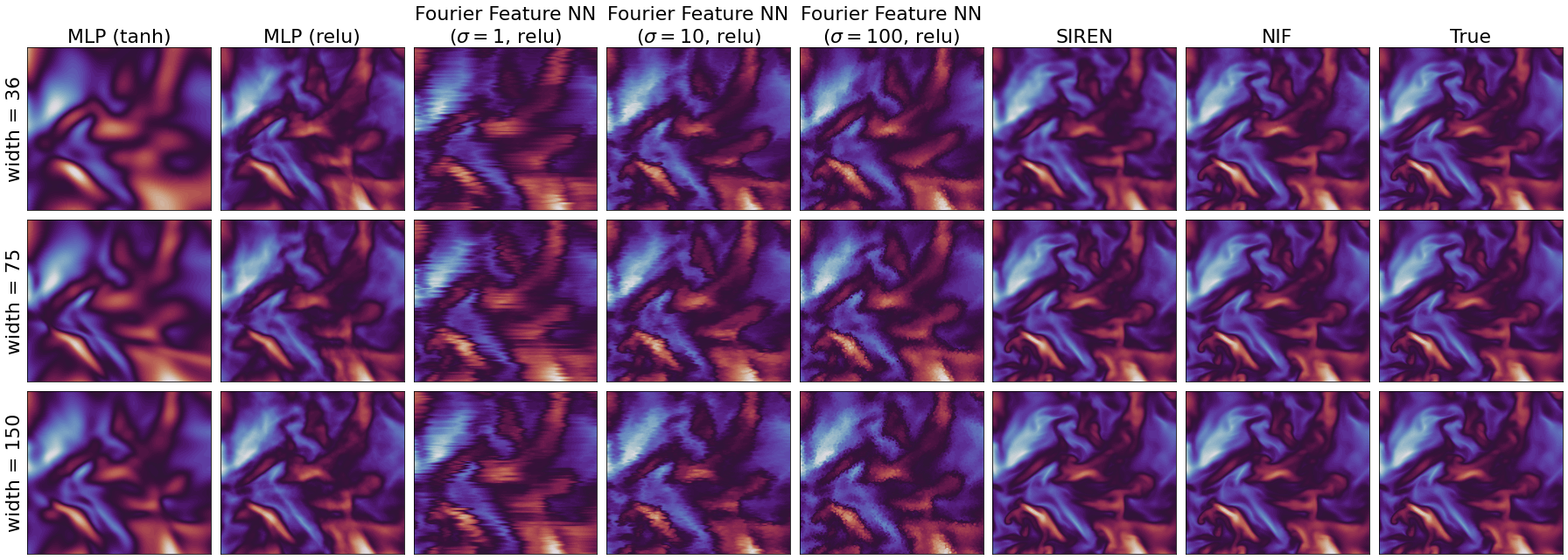}
\caption{Comparison between NIF and other popular frameworks for reconstructing the $x$-velocity of a certain slice in 3D homogeneous isotropic turbulence. Each column corresponds to the reconstruction result for a certain model except last column which is the ground truth. Each row corresponds to the same network width associated with spatial input. }
\label{fig:2dturb}
\end{figure}

Finally, we compare NIF against popular frameworks \emph{under the same computational complexity for spatial query}, i.e., the same network width associated with spatial input, for the task of reconstructing a toy 2D video of turbulence. \textcolor{black}{ \Cref{fig:2dturb} qualitatively shows the comparison of our framework against standard MLPs, Fourier Features Networks~\citep{tancik2020fourier}, and SIREN~\citep{sitzmann2020implicit} on a toy time-varying 2D dataset containing a slice of 3D homogeneous isotropic turbulence. From \cref{tab:2dturb}, we confirm that NIF performs the best among all of the frameworks especially when the network width is limited (e.g., 36, 75) while comparable to vanilla SIREN when the network width becomes larger.}

\begin{table}[bthp]
\centering
\scriptsize
\caption{Normalized error between prediction and ground truth of the 2D time-varying $x-$velocity from a turbulent flow. Frobenius norm of the ground truth is the normalization factor. The configurations for the same row share the same network width, which approximately determines the computational complexity at the inference stage. We found Fourier NN requires a non-trivial tuning for the frequency $\sigma$ and it doesn't outperform SIREN and NIF. Our framework NIF performs the best for small to middle-range network width while comparable to SIREN when network width reaches 150.}
\label{tab:2dturb}
\vspace{1em}
\begin{tabular}{cccccccc}
\makecell[c]{\textbf{Network}\\ \textbf{width}}   &   \makecell[c]{\textbf{MLP} \\\textbf{(tanh)}} &   \makecell[c]{\textbf{MLP}\\\textbf{(relu)}} &  \makecell[c]{ \textbf{Fourier NN} \\ \textbf{$(\sigma=1)$ }} & \makecell[c]{ \textbf{Fourier NN} \\ \textbf{ $(\sigma=10)$} }  &\makecell[c]{ \textbf{Fourier NN} \\ \textbf{$(\sigma=100)$} }  & \textbf{SIREN} & \makecell[c]{\textbf{NIF}\\ (Ours)} \\ 
\midrule
36  & 0.357    &    0.196 &   0.391                &    0.265                    &    0.263                     & 0.121            & \textbf{0.071}   \\
75  & 0.290    &    0.172 &   0.361                &    0.260                    &    0.241 &  0.040 &  \textbf{0.022} \\
150 &  0.236 &    0.144 &      0.353 &     0.285 &      0.248 &  \textbf{0.0116} &  0.0131   \\
\bottomrule
\vspace{1em}
\end{tabular}
\end{table}

\subsection{Modal analysis on adaptive mesh data: flow over cylinder}
\label{sec:app_modal_analysis}

Next, we test this NIF-based framework to learn a linear subspace for a classical modal analysis problem in fluid mechanics: vortex shedding behind a cylinder.
As shown in \cref{fig:linear_subspace_cyd}, the simulation is performed with AMR,  which is frequently seen in computational fluid dynamics for efficient simulation and data-storage for a wide range of flows containing moving sharp gradients (e.g., unsteady shocks/flames/vortices). Here we first collect 100 snapshots of a spatial field consisting of two velocity component $u$ and $v$, together with $x$, $y$ coordinate in each cell and time $t$. Then, we follow \cref{eq:nif_linss} to extract a rank-10 linear subspace from the spatio-temporal field. NIF with SIREN is employed in both ParameterNet and ShapeNet. Details of the data preparation and the reconstruction performance are given in \cref{apdx:cyd}. Given the learned latent 10-dimensional time series $\zeta_1 = a_{\textrm{MLP},1}(t; \mathbf{\Theta}),\ldots,\zeta_{10}=a_{\textrm{MLP},10}(t; \mathbf{\Theta})$ as shown in \cref{fig:linear_subspace_cyd}, we perform a standard dynamic mode decomposition (DMD) \citep{schmid2022dynamic} to extract isolated spatio-temporal modes with distinct frequencies shown. The DMD mode shapes in \cref{fig:linear_subspace_cyd} agree qualitatively well with other existing studies \citep{pan2020sparsity,chen2012variants}. Note that the latent representation $a_{\textrm{MLP},i}$ contains time $t$ and spatial functions $\bm{\phi}_{\textrm{MLP},i}$ contains $\mathbf{x}$ as input arguments. Thus, at the postprocessing stage, one can easily evaluate these functions above at any time $t$ and/or spatial location $\mathbf{x}$ for any resolution one desires.

\subsection{Data-driven sparse reconstruction of sea surface temperature}
\label{sec:app_sparse_sensing}

\begin{figure}[htbp]
\centering
\includegraphics[width=\textwidth]{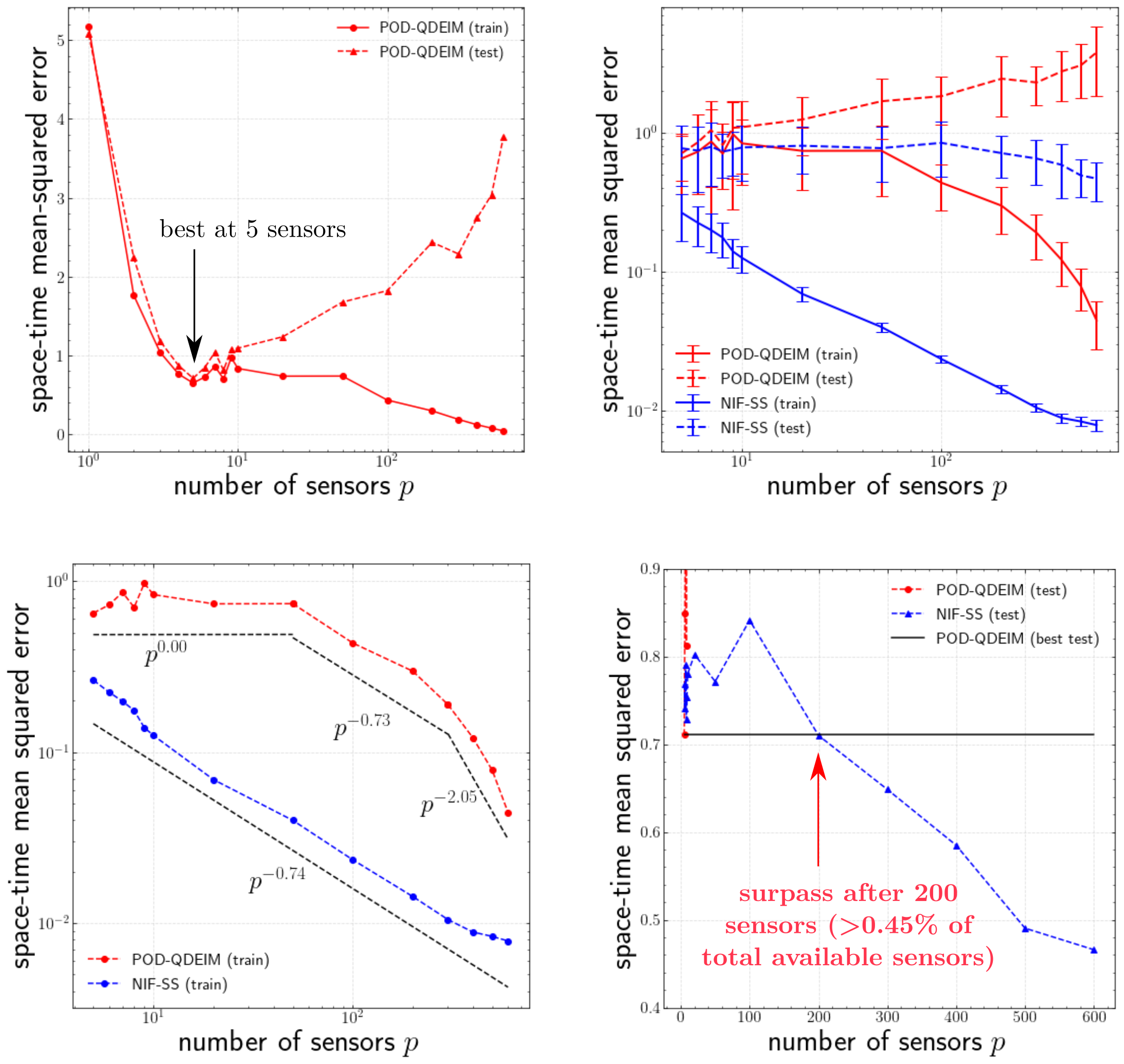}
\caption{Comparison between NIF-based sparse sensing and POD-QDEIM. \textbf{Top left:} training and testing error of POD-QDEIM with a complete parameter sweep for the number of sensors from $1$ to $600$. \textbf{Top right:} mean-squared error with standard deviation for varying number of sensors from 5 to 600. Error bar represents plus and minus one temporally standard deviation of spatially mean square error. \textbf{Bottom left:} approximated decay rate of training error with respect to increasing number of sensors $p$ (POD-QDEIM versus our NIF-based framework). \textbf{Bottom right:} comparison of testing error among NIF-SS, POD-QDEIM and best POD-QDEIM model. }
\label{fig:ss_mse_compare}
\end{figure}

As shown in \cref{fig:nif_sparse_sensing}, we apply the above NIF-based framework to reconstruct and predict sea surface temperature data \citep{reynolds2002improved} given sparse localized measurements. In order to compare with the state-of-the-art, we use a similar setup from \cite{manohar2018data}. 
We take snapshots from 1990 to 2006 as training data and that of the next 15 years, until 2021, as testing data. As mentioned in the previous subsection, we first use POD-QDEIM on the training data to find the best-$p$ sensor locations on the sea. Besides, as shown in the top left of \cref{fig:ss_mse_compare}, we empirically find POD-QDEIM performs the best at surprisingly low 5 sensors. In order to perform an exhaustive evaluation on NIF-based framework, we vary $p$ from 5 to 600. Due to the appearance of multi-scale structure in sea surface temperature, we use NIF with SIREN in \cref{sec:ct}. For $p=5$ to $p=60$, we take 2 ResNet-like blocks in ShapeNet with width 60 and 2 blocks in ParameterNet with width 60. For $p > 60$, we take 2 blocks in ShapeNet still with width 60 and 2 blocks in ParameterNet with width $p$. For all cases, we fix $n_p=p$ in analogous to the equality between rank and number of sensors in POD-QDEIM. We use Adam optimizer for mini-batch training with learning rate as 1e-5 and batch size as 7000 for 150 epochs. Details of data preparation are in \cref{apdx:sst}.

To further evaluate our framework, we make a side-by-side comparison with the state-of-the-art POD-QDIEM in both reconstruction and prediction of sea surface temperature using the same information from the best-$p$ sparse sensor locations from POD-QDEIM. As displayed in the top right of \cref{fig:ss_mse_compare}, the space-time mean-squared error on training data of both NIF-based framework and POD-QDEIM decrease as the number of sensors increase while that of our framework decays much more quickly than that of POD-QDEIM. The approximated decay rate is shown in the bottom left of \cref{fig:ss_mse_compare}. We find that our NIF-based framework shows a consistent decay rate $-0.74$ as the number of sensors $p$ increases. On the other hand, POD-QDEIM struggle to decrease training error only after $p>50$ with a similar decay rate of $-0.73$ with a faster rate of $-2.05$ after $p>300$. Also, it is interesting to note that as more and more sensors are used, the POD-QDEIM generalization is worse and worse while our framework in general performs better and better. As shown in the bottom right of \cref{fig:ss_mse_compare}, given the number of sensors $p$ considered in this work, our NIF-based model surpass the best POD-QDEIM model after using 200 sensors, which corresponds to using more than $0.45\%$ of all possible sensor locations on the sea. Finally, as mentioned before, the most generalizable model of POD-QDEIM is using 5 sensors which results in a space-time M.S.E as 0.71 (additional parameter sweeps are shown in the top left of \cref{fig:ss_mse_compare}). While the model of our NIF-based framework with smallest testing error takes 600 sensors and results in a space-time M.S.E as 0.46, which is 34\% smaller than the best model of POD-QDEIM.

Apart from comparing space-time mean squared error, we also compute a standard deviation of spatially mean squared error along time axis as an indicator for robustness of model performance (see error bars in \cref{fig:ss_mse_compare}). Note that the $y$ axis of the top right figure in \cref{fig:ss_mse_compare} is in log scale. Therefore, for the range of number of sensors considered, training error bar of our framework is significantly smaller than that of POD-QDEIM, which indicates our framework can faithfully reconstruct the training data with higher confidence. This is particularly important for hyper-reduction in projection-based ROMs~\citep{carlberg2011efficient}. The testing error bar of our framework is also smaller than that of POD-QDEIM, which means that our framework has higher robustness in predicting unseen data as well. Additional discussions are in \cref{apdx:sparse_rec}.

\section{Conclusions}
\label{sec:conclusion}

High-dimensional spatial complexity is a major bottleneck of computational and storage costs in many physical science and engineering fields where the physics relies on a set of complex partial differential equations (PDEs). Existing frameworks, such as SVD and CAE, suffer from various challenges arising from complex data structures in real world scientific computing. In this work, a mesh-agnostic representation for parametric spatial-temporal datasets, called \emph{neural implicit flow} (NIF), is proposed. The key feature of NIF is its ability to separate spatial complexity from other factors such as temporal and parametric complexity, which naturally follows the philosophy of manifold-based learning of PDEs. Compared to existing SVD and CAE frameworks, of which the performance is either restricted by unnecessary projection, linearity, intractable memory scaling for large-scale 3D dataset, or inefficiency for adaptive meshes, NIF enables scalable mesh-agnostic nonlinear dimensionality reduction with improved performance. As a comparison, \cref{table:compare} shows a summary of the capabilities and challenges of SVD, CAE, and NIF. Specifically, we demonstrate the advantages of NIF in terms of four derived \textit{mesh-agnostic} frameworks: data-fit surrogate modeling for parametric PDEs, compressed representation of multi-scale spatial-temporal data, learning linear representations, and nonlinear sparse sensing. To the best of our knowledge, nonlinear dimensionality reduction of 3D turbulence with over 2 million cells and complex multi-scale dynamics with an AMR solver is achieved for the first time.

\subsection{Disadvantages}
\textcolor{black}{
\paragraph{Intrinsic data complexity:} For spatially complex data, one requires a correspondingly large ShapeNet to accommodate the spatial structures. Larger networks require longer time to converge. As a result, one has to manually decide the size of ShapeNet and ParameterNet for specific problems, whereas SVD and CAE are much easier to configure.  
}
\textcolor{black}{
\paragraph{Long training time:} Unlike SVD and CAE where best practices are well established~\citep{maulik2021pyparsvd,he2019bag}, training a hypernetwork of a deep neural network still requires some trial and error. A typical SVD runtime takes a few seconds to minutes. Training CAE usually takes 0.5hrs to arrive at a decent accuracy. While NIF usually takes above an hour to days depending on the data complexity, which in turns affects the size of model. First, because the input of NIF is pointwise, the total number of training data can be huge for 3D datasets. Second, in order to connect the two networks, there are expensive tensor operations between 1D and a 2D tensor (output reshaped from ParameterNet). Note that both of the tensors are \emph{batch dependent} in NIF, which can decrease cache performance. For example in the 3D turbulence case, it takes up to 1.3 hours to process one single epoch and almost 5 days on a 16GB GPU to obtain a good accuracy. 
}
\textcolor{black}{
\paragraph{Memory intensive:} The use of hypernetwork leads to more complex network topology than the standard feedforward neural net with a comparable model parameter size. As a result, it creates larger memory footprints, which unfortunately limits the maximal batch size and leads to longer training time. 
}
\textcolor{black}{
\paragraph{Lack of invariance:} Despite of some promising results~\citep{wang2022mosaic}, it is generally more challenging to embed invariance into coordinate-input neural networks than graph-based approaches~\citep{li2020fourier}. Such lack of invariance may worsen the generalization of data-fit regression especially when the amount of training data is small. 
}
\subsection{Advantages}
\textcolor{black}{
\paragraph{Intrinsic data complexity:} On the flip side, model complexity does not scale with the ``superficial'' complexity of the data, e.g., the number of mesh points. Finer mesh points only lead to more training data while model complexity can keep the same. Meanwhile, mesh-based models (e.g., CAE) still suffer from growing model complexity as the number of mesh points increases.
}
\textcolor{black}{
\paragraph{Heterogeneous data sources:} Since it is mesh-agnostic in nature, it becomes very convenient to fuse information from different data sources. For example, PIV data are mostly on a perfect uniform Cartesian mesh, whereas CFD data are commonly on a well designed body-fitted mesh. Sometimes different types of CFD solver would use different mesh, e.g., multi-level Cartesian mesh for LBM/IBM~\citep{taira2007immersed}. 
}
\textcolor{black}{
\paragraph{Well-established manifold-based ROM:} Thanks to the decoupling of spatial complexity, we have a direct analogy of SVD but with a mesh-agnositic, nonlinear and scalable version for 3D datasets. It is straightforward to extend the established ROM frameworks~\citep{carlberg2011efficient,bruna2022neural}, modal analysis~\citep{schmid2022dynamic,taira2020modal,rowley2009spectral} to more realistic and mesh-complex datasets. 
}
\textcolor{black}{
\paragraph{Efficient spatial query:} Postprocessing of PDE data, e.g., turbulence \citep{li2008public},  often involves intensive spatial query than temporal or other parametric query. Our design of NIF leads to a compact network for spatial complexity, which improves the efficiency for massive spatial query on either the point value or the spatial derivative. 
}
\section{Prospects}

NIF has the potential to extend existing data-driven modeling paradigms based on SVD and CAE to the real world of scientific computing, where raw spatial-temporal data can be three dimensional, large-scale, and stored on arbitrary adaptive meshes. High-fidelity large-scale data from modern scientific computing codes~\citep{almgren2010castro,nonaka2012deferred} can be reduced to effective low dimensional spaces, where existing modal analysis~\citep{taira2020modal,towne2018spectral,mckeon2010critical} data-driven flow control~\citep{duriez2017machine} and surrogate-based modeling/optimization~\citep{koziel2013surrogate} can be performed seamlessly on the arbitrary adaptive mesh. However, as is typical for nonlinear dimensionality reduction methods (e.g., CAE), the latent representation is more difficult to interpret than its linear counter part (e.g., SVD), and each time the latent representation can be different. This raises new questions on how to guide the latent representation to be interpretable for experts. \textcolor{black}{Another exciting direction is to construct projection-based ROMs in such a latent space by minimizing the residual of \emph{known} governing equations. This has been demonstrated very recently on the material point method \citep{chen2021model} and for high-dimensional PDE with active learning \citep{bruna2022neural}.}
Besides dimensionality reduction, NIF can be used to design efficient ``decoders'' for transferring large spatial scientific data. It essentially trades reduction in storage/transfer and ease in data fusion of heterogeneous meshes with off-line training a NIF model and additional function calls on the ShapeNet.

\begin{table}
\centering
\footnotesize
\caption{Capabilities and challenges of representation learning: SVD, CAE, and Neural Implicit Flow.}
\label{table:compare}
\vspace{1em}
\begin{tabular}{llll}
\textbf{Property/Model} & \textbf{SVD} & \textbf{CAE} & \textbf{Neural Implicit Flow} \\ 
\midrule 
\vspace{1em}
\makecell[l]{Resolution\\ \quad \\ \quad }  
&
\makecell[l]{\textcolor{OliveGreen}{strong}\\Convergence to \\discrete data}
&
\makecell[l]{\textcolor{red}{weak}\\Requires uniform mesh\\ \quad }
&
\makecell[l]{\textcolor{OliveGreen}{strong}\\ Continuous field \\ \quad } 
\\ \vspace{1em}
\makecell[l]{Variable geometry} 
&
\makecell[l]{\textcolor{OliveGreen}{{strong}}}   
&
\makecell[l]{\textcolor{red}{weak}}
&
\makecell[l]{\textcolor{OliveGreen}{strong}\\} \\ \vspace{1em}
\makecell[l]{Scalablity \\ \quad \\ \quad \\ \quad } 
&
\makecell[l]{\textcolor{OliveGreen}{{strong}} \\ Efficient randomized \\ SVD available \\ \quad }   
&
\makecell[l]{\textcolor{BurntOrange}{fair}/\textcolor{red}{weak} \\ Affordable in 2D \\ Resolution restricted in 3D \\ \quad }
&
\makecell[l]{\textcolor{OliveGreen}{strong}\\ Point-wise mini-batch \\ training; Number of \\ parameters  required \\ only scale with\\ intrinsic complexity} \\ \vspace{1em}
\makecell[l]{Parametric/temporal\\variation of domain\\ discretization}   
&
\makecell[l]{\textcolor{red}{weak}\\Requires the same\\mesh throughout}
&
\makecell[l]{\textcolor{red}{weak}\\Requires the same\\uniform mesh throughout}
&
\makecell[l]{\textcolor{OliveGreen}{strong} \\ arbitrary mesh \\ \quad } \\ \vspace{1em}
\makecell[l]{Training easiness} 
&
\textcolor{OliveGreen}{strong}
&
\textcolor{BurntOrange}{fair}/\textcolor{OliveGreen}{strong}
&
\makecell[l]{\textcolor{BurntOrange}{fair}} \\ \vspace{1em}
\makecell[l]{Interpretability of \\ representation \\ \quad } 
&
\makecell[l]{\textcolor{OliveGreen}{strong}\\ \quad \\ \quad }
&
\makecell[l]{{\textcolor{red}{weak}} \\ \quad \\ \quad }
&
\makecell[l]{\textcolor{BurntOrange}{fair} \\ last-layer parameterization \\ learns linear subspace } \\ \vspace{1em}
\makecell[l]{Expressiveness \\ \quad \\ \quad } 
&
\makecell[l]{\textcolor{red}{weak}\\Linear, not ideal \\ for advection \\dominated flows}
&
\makecell[l]{\textcolor{BurntOrange}{fair}\\ Nonlinear but cannot capture\\ multi-scale efficiently \\ \quad}
&
\makecell[l]{\textcolor{OliveGreen}{strong}\\ Nonlinear with multi-scale \\capability \quad} \\
\bottomrule 
\end{tabular}
\end{table}


\acks{The authors thank Kevin Carlberg, Karthik Duraisamy, Zongyi Li, and Lu Lu for valuable discussions. 
The authors acknowledge funding support from the Air Force Office of Scientific Research (AFOSR FA9550-19-1-0386). 
The authors acknowledge funding from the National Science Foundation AI Institute in Dynamic Systems grant number 2112085.
We also appreciate the generous support from Prof. Karthik Duraisamy on using the following computing resources, which were provided by the NSF via the grant ``MRI: Acquisition of ConFlux, A Novel Platform for Data-Driven Computational Physics''. 
This work also used the Extreme Science and Engineering Discovery Environment (XSEDE)~\citep{xsede}, which is supported by National Science Foundation grant number ACI-1548562. Specifically, it used the Bridges-2 system, which is supported by NSF award number ACI-1928147, at the Pittsburgh Supercomputing Center (PSC).}


\newpage

\pagebreak
\appendix

\section{POD and CAE}

\begin{figure}[t]
\centering
\includegraphics[width=\textwidth]{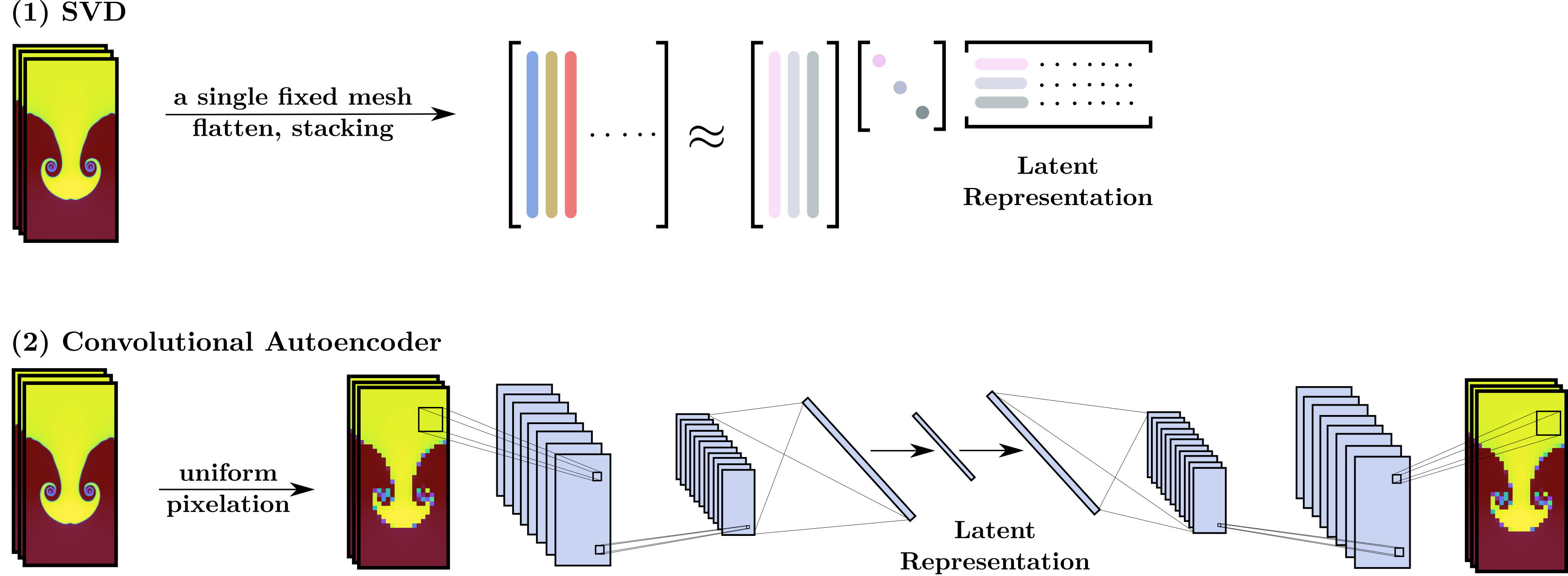}
\caption{SVD and CAE for dimensionality reduction of spatio-temporal field from PDE. (1) SVD relies on the assumption that spatio-temporal field is sampled from a single fixed mesh for all parameters across all time. Here the latent representation is formed by the right singular vectors. (2) CAE treats spatio-temporal fields as images by pre-processing with uniform pixelation. Latent space is the flattened vector after several convolution and pooling operations.}
\label{fig:apdx_cae_pod}
\end{figure}

As shown in \cref{fig:apdx_cae_pod}, state-of-the-art methods for dimensionality reduction of parametric spatial temporal data rely on SVD and CAE. Dimensionality via SVD requires the data on a single fixed mesh. First, it flattens the spatial data into a long column vector. Second, these column vectors are stacked over time. Finally, SVD is performed on such stacked matrix and the right singular vectors are the latent representation. CAE treated the flowfield as image. First, one performs a uniform pixelation. Second, one feed the processed images into convolutional autoencoders. The corresponding latent representation is formed by the bottleneck layer in \cref{fig:apdx_cae_pod}. 

\section{Data preparation}

\subsection{1D parametric K-S}
\label{apdx:ks}

\begin{figure}[t]
\centering
\includegraphics[width=\textwidth]{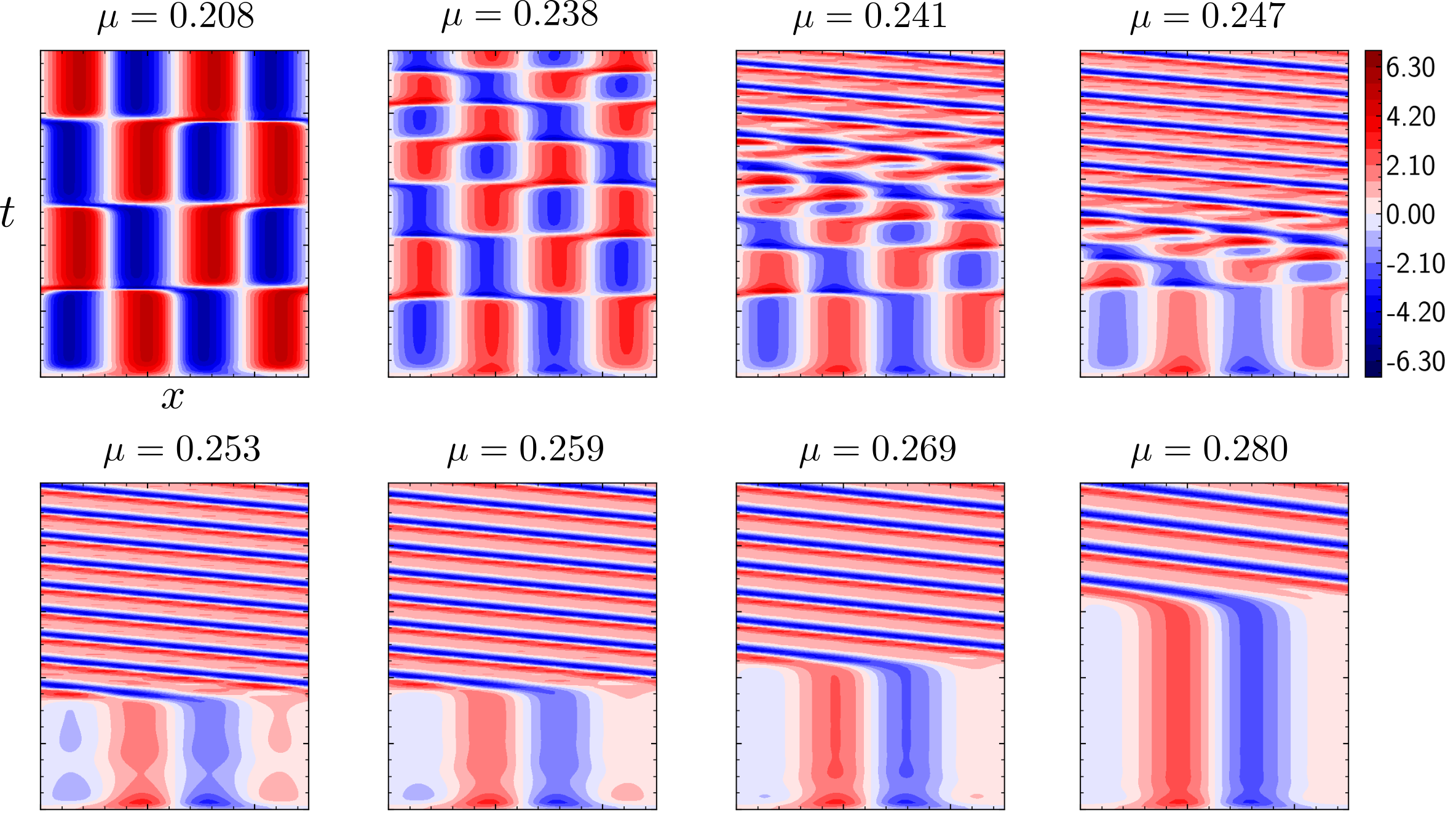}
\caption{Parametric variation of spatial temporal field generated from Kuramoto Sivashinsky system in $x-t$ at different $\mu$.}
\label{fig:visualize_ks_data}
\end{figure}

Given $x\in [0,2\pi]$, $t\in [0,100]$, we use ETD-RK4 \citep{kassam2005fourth} method to solve 1D Kuramoto-Sivashinsky equation in \cref{eq:ks}, with periodic boundary condition $u(0,t)=u(2\pi,0)$ and a fixed initial condition $u(x,0)=\sin(x)$. Spatial discretization is performed with 1024 points. Time integration $dt = 10^{-3}$. The raw dataset has space-time resolution as $1024 \times 10000$. We subsample 4 times in space and 1000 times in time, i.e., $256 \times 100$. Note that K-S has a rich dynamics when parameter changes \citep{papageorgiou1991route}. In order to make the problem well-posed, we choose a regime $\mu \in [0.2,0.28]$ where the system is not chaotic. Parametric variation of the solution in $x-t$ is displayed in \cref{fig:visualize_ks_data}. Training data is prepared with 20 uniform sampling with $[0.2,0.28]$, ending up with $20 \times 256 \times 100 = 0.512 \times 10^{6}$ data points. Testing data is prepared with $40$ uniform samples within $[0.2,0.28]$, which leads to $40 \times 256 \times 100 = 1.024 \times 10^{6}$ data points. Still, each data point has 4 components $\mu,t,x,u$. Each component is normalized with zero mean and unit standard deviation. 

\subsection{2D Rayleigh-Taylor}
\label{apdx:rt}

Rayleigh-Taylor instability happens when a heavy fluid floats above a light fluid in a gravitational field. The interface between heavy and light fluid is unstable and ``mushroom''-like multi-scale structure will grow as small-scale structures grows much faster than large-scale structures. In order to efficiently simulate the system, adaptive mesh refinement is necessary. The goal of dimensionality reduction of parametric fluid system is to reduce the spatial complexity so that one may perform modeling, design or control in the reduced coordinate efficiently. However, without given the range of system parameters in the data, it is often difficult to make statement on the true dimensionality of a flow. In order to simplify our discussion, we choose only to simulate a single realization of R-T with fixed system parameter where the dimensionality can be reduced to one. 

Using CASTRO, we consider one level of AMR with an initial Cartesian grid of 256 points along vertical axis and 128 points along horizontal axis during the simulation, which efficiently captures the evolution of vortex generation and advection of density fronts. AMR refinement ratio is 2 in each direction. We regrid at every 2 steps. The simulation domain is a rectangular with 0.5 long in $x$-axis and 1.0 in $y$-axis. Top and bottom boundary condition is slip wall while left and right are periodic. Time span of the simulation is from 0 to 3.5. The density for heavy fluid is 2 while that for the light fluid is 1. They are separated by a horizontally trigonometric and vertically tanh interface, which is a perturbation to the unstable system. The exact expression follows the equation (46-47) in the original CASTRO paper \citep{almgren2010castro}. 

We save the simulation data at every 10 steps. Training data starts from 83rd to 249th snapshot (corresponding to time $t$ from 1.004 to 2.879) with skipping on the even snapshot: 83, 85, \ldots, 247, 249. Testing data starts from 86th to 248th snapshot with an incremental of six: 86, 92, \ldots, 242, 248. Training and testing data are not overlapping but they have no distribution shift. In total, training data has 84 snapshots and testing data as 28 snapshot. The number of adaptive mesh points ranges from 40,976 to 72,753 across time. For POD and CAE, original data on adaptive mesh is projected onto three different Cartesian grids: $32\times 64$, $64\times 128$, $128\times 256$ using the filter \texttt{ResampleToImage} in \texttt{ParaView}~\citep{ahrens2005paraview}. Finally, original data is also projected onto a very fine Cartesian mesh $256\times 512$ as ``ground truth'' in order to make quantitative comparison for predictions from different models.

\begin{figure}[t]
\centering
\includegraphics[width=0.8\textwidth]{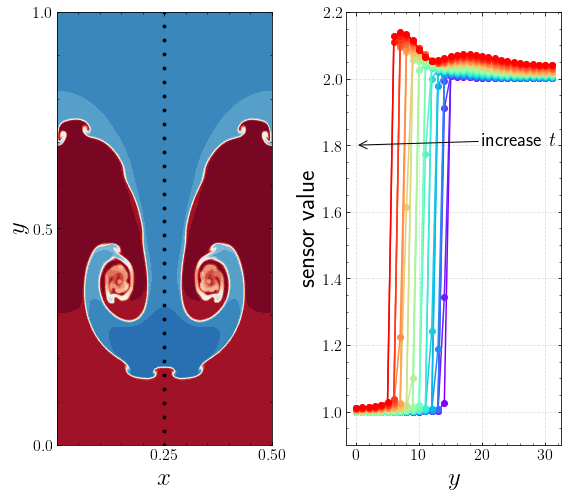}
\caption{Illustration of sensor measurements for NIF-autoencoder in the R-T instability. Left: 32 sensor locations are distributed evenly on the middle vertical axis. Right: all 32 sensor values for different time $t$ in the testing data.}
\label{fig:rt_sensors}
\end{figure}

As for NIF, firstly, note that we use a feedforward neural network encoder with sparse sensor as input. As shown in the left of \cref{fig:rt_sensors}, we collect measurements from 32 equally spaced sensor along the vertical axis in the middle of the flowfield. At test time, input data that feed to the encoder is displayed right of \cref{fig:rt_sensors}, where we can see a clear density front moving towards the bottom. Secondly, since NIF takes pointwise data, each data point is a 35-dimensional row vector. The first 32 are sensor values at $t$ while the rest three are $x,y$ and density at that location $u(x,y)$. Because we need to go through all grid points in the adaptive mesh at time $t$, there will be as many data points with repeated 32 columns as the total number of points in the adaptive mesh at time $t$. In fact, we end up with 4,190,928 rows with 35 columns for the training data. Thirdly, since we are using NIF with SIREN in our ShapeNet, we normalize $x,y$ between -1 and 1 while the rest 33 columns are normalized with zero mean and unit standard deviation. 

\subsection{2D Cylinder Flow with ${Re}=123$}
\label{apdx:cyd}

\begin{figure}[t]
\centering
\includegraphics[width=0.85\textwidth]{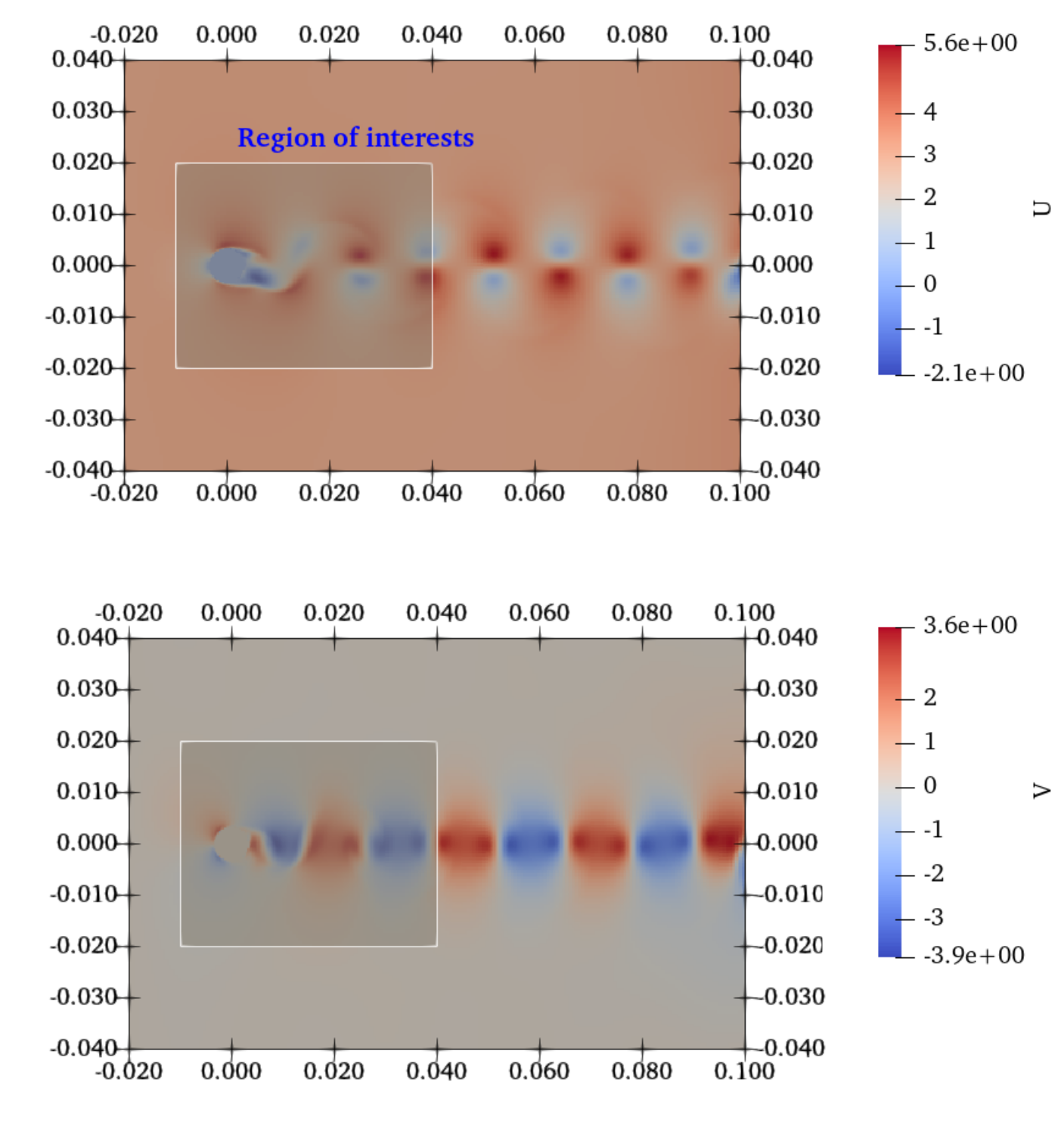}
\caption{Configuration and mesh of CFD simulation on 2D cylinder flow at $Re \approx 123$ with AMR. Contour of magnitude of vorticity is shown. Domain of interests is marked in light gray with white edges.}
\label{fig:config_cyd}
\end{figure}

We generated the data using \textsl{PeleLM} \citep{nonaka2012deferred} to solve a incompressible N-S. The cylinder geometry is represented using embedding boundary method. The setup is given in our the Github repo. We set the cylinder centered at the original. The computational domain is $x$ (streamwise) direction from -0.02 to 0.1 while $y$ direction from -0.04 to 0.04. The boundary condition is set as inflow: Dirichlet, outflow Neumann. Side: periodic. We set $U_\infty = 3$ m/s, viscosity $\mu_\infty = 2\times 10^{-4}$ Pa$\cdot$s, $\rho_\infty=1.175$ kg/$\textrm{m}^3$. The radius of cylinder $r=0.0035 $ m. Thus, Reynolds number is $Re = {2U_{\infty}r}/{\mu_\infty}\approx 123$. The simulation is performed from $t=0$ s to $t=2$ s with sampling $\Delta t= 0.0005$ s. The flow is initialized with uniform streamwise flow $U_\infty$ superimposed with a side velocity in order to break the symmetry so that the flow can quickly fall on to the limit cycle. To remove the initial transient effect, we collect the last 100 snapshots. The AMR is based on vorticity where we enable a finer mesh with half of the grid size if the magnitude of local vorticity monitored exceed 3000. Overall, we collect data (we can easily collect AMR data pointwise using ParaView) after the system falls on to the limit cycle and sampled the last 100 snapshots. Moreover, in order to remove the effect from the exit boundary and only focus on the region of interests, we only sample cell-centered data within the domain of interests with $x\in[-0.01, 0.04]$ m and $y\in[-0.02, 0.02]$ m as shown in \cref{fig:config_cyd}. Since we are using NIF with SIREN in \cref{apdx:nif_sine}, we normalize the $t,x,y$ into uniform distribution between -1 and 1. For cell area $\Delta \mathbf{x}$, we scale it with a factor of $10^6$ since we are using single precision in Tensorflow. For output velocity $u,v$, we first normalize them into zero mean. Next we normalize $u$ into unit variance and normalize $v$ with the same factor. 
Finally, we arrange the collected pointwise data into a big matrix, with 966,514 rows and 6 columns, 
$$
\begin{bmatrix}
\widetilde{t}_1 & \widetilde{x}_1 & \widetilde{y}_1 & \widetilde{u}_1 & \widetilde{v}_1 & \widetilde{\Delta\mathbf{x}}_1 \\
\vdots & \vdots & \vdots & \vdots  & \vdots & \vdots\\ 
\widetilde{t}_{966514} & \widetilde{x}_{966514} & \widetilde{y}_{966514} & \widetilde{u}_{966514} & \widetilde{v}_{966514} & \widetilde{\Delta\mathbf{x}}_{966514}
\end{bmatrix},
$$
where wide tilde means normalized variable.

\subsection{3D Forced Homogeneous Isotropic Turbulence with ${Re}_{\lambda}=433$}
\label{apdx:3dhit}

We use the \textit{forced isotropic turbulence} dataset from JHU Turbulence dataset \citep{li2008public} with Taylor-scale Reynolds number ${Re}_{\lambda}$ around 433. The simulation is computed by solving 3D incompressible Navier-Stokes equation with pseudo-spectral method. To sustain the turbulence, energy is injected into the large scales. After the system reaches the equilibrium state, the original dataset consists snapshots collected at every 0.02 nondimensionalized time. Here we collect $1,4,7,10,\ldots, 58$-th snapshot, in total 20 snapshots. 
Then we slice a block with $128^3$ resolution from the original HIT with the 20 snapshots, which is $1024^3$. Since we are using NIF with SIREN in \cref{apdx:nif_sine}, we normalize the range of $t,x,y$ so that the input signal is uniformly distributed between -1 and 1. For target velocity $u,v,w$, we simply normalize them into zero mean and unit variance. 

\subsection{Sea Surface Temperature}
\label{apdx:sst}

We obtain the weekly averaged sea surface temperature data since 1990 to present from NOAA website \footnote{\url{https://downloads.psl.noaa.gov/Datasets/noaa.oisst.v2/sst.wkmean.1990-present.nc}}. At the time when we worked on this problem, there is in total 1642 snapshots (1990-2021) with each snapshot corresponds to a weekly averaged temperature field. Each temperature snapshot contains 180 uniformly spaced latitude and 360 uniformly spaced longitude coordinates. It should be noted coordinates correspond to land should be masked \footnote{\url{https://downloads.psl.noaa.gov/Datasets/noaa.oisst.v2/lsmask.nc}}. Thus, each masked snapshot contains 44219 points. We use temperature field of the first 832 weeks (16 years) as training data with a total number of  36,790,208 data points. Each data point is a $n_s+3$-dimensional row vector, where $n_s$ is the number of sensors and 3 corresponds to $x,y,T$. Locations of $n_s$ sensors are obtained via POD-QDEIM or equivalently data-driven sensor placement in \citep{manohar2018data}.  Temperature field in the rest of time are testing data. Still, spatial coordinate $x$ and $y$ is normalized between -1 and 1. Target temperature is normalized into zero mean and unit variance.

\section{Additional discussions}

\subsection{Data-fit surrogate modeling for 1D Kuramoto-Sivashinsky}
\label{apdx:datafit}

\begin{figure}[t]
\centering
\includegraphics[width=\textwidth]{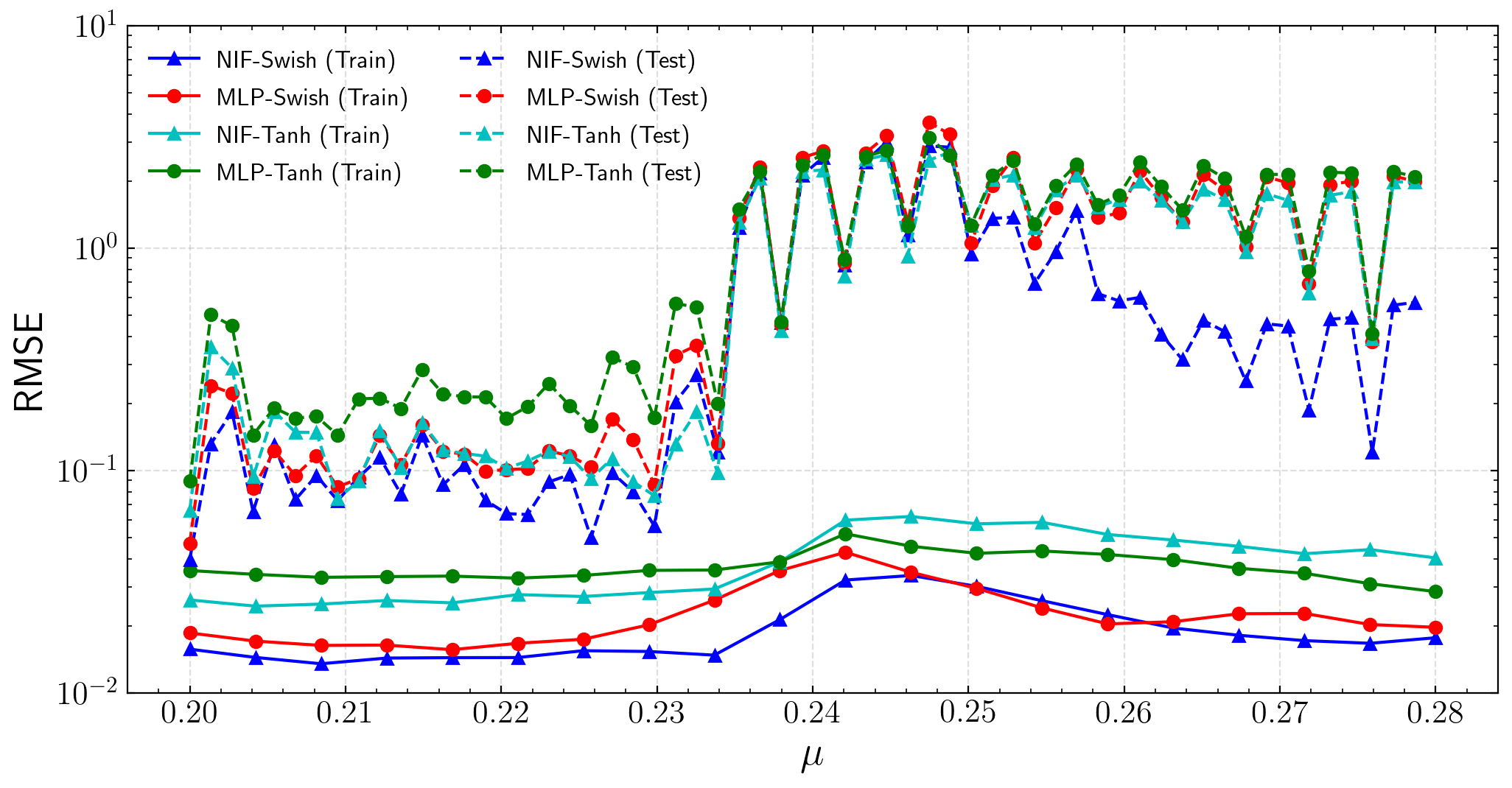}
\caption{Comparison among four model configurations (NIF-Swish, MLP-Swish, NIF-tanh, MLP-tanh) in terms of RMSE on the dataset of 1D parametric KS system with $0.2<\mu<0.28$. }
\label{fig:ks_pm}
\end{figure}

Recall that the advantages of NIF have been demonstrated in \cref{sec:app_data-fit-1d-ks}. In order to further compare the model performance, we further compute the RMSE for each parameter $\mu$ for four configurations: (1) NIF-Swish, (2) MLP-Swish, (3) NIF-tanh, (4) MLP-tanh. As displayed in \cref{fig:ks_pm}, NIF with Swish generalizes better than other configurations especially when $\mu > 0.25$. As seen in \cref{apdx:ks}, such parameter regime corresponds to the traveling waves solutions with a transition time increasing with $\mu$. Meanwhile, NIF-tanh generalizes slightly better tanh MLP-tanh but similar with MLP-Swish. Hence, we only compare NIF-Swish with MLP-Swish in the rest of the paper.

\begin{figure}[t]
\centering
\includegraphics[width=0.68\textwidth]{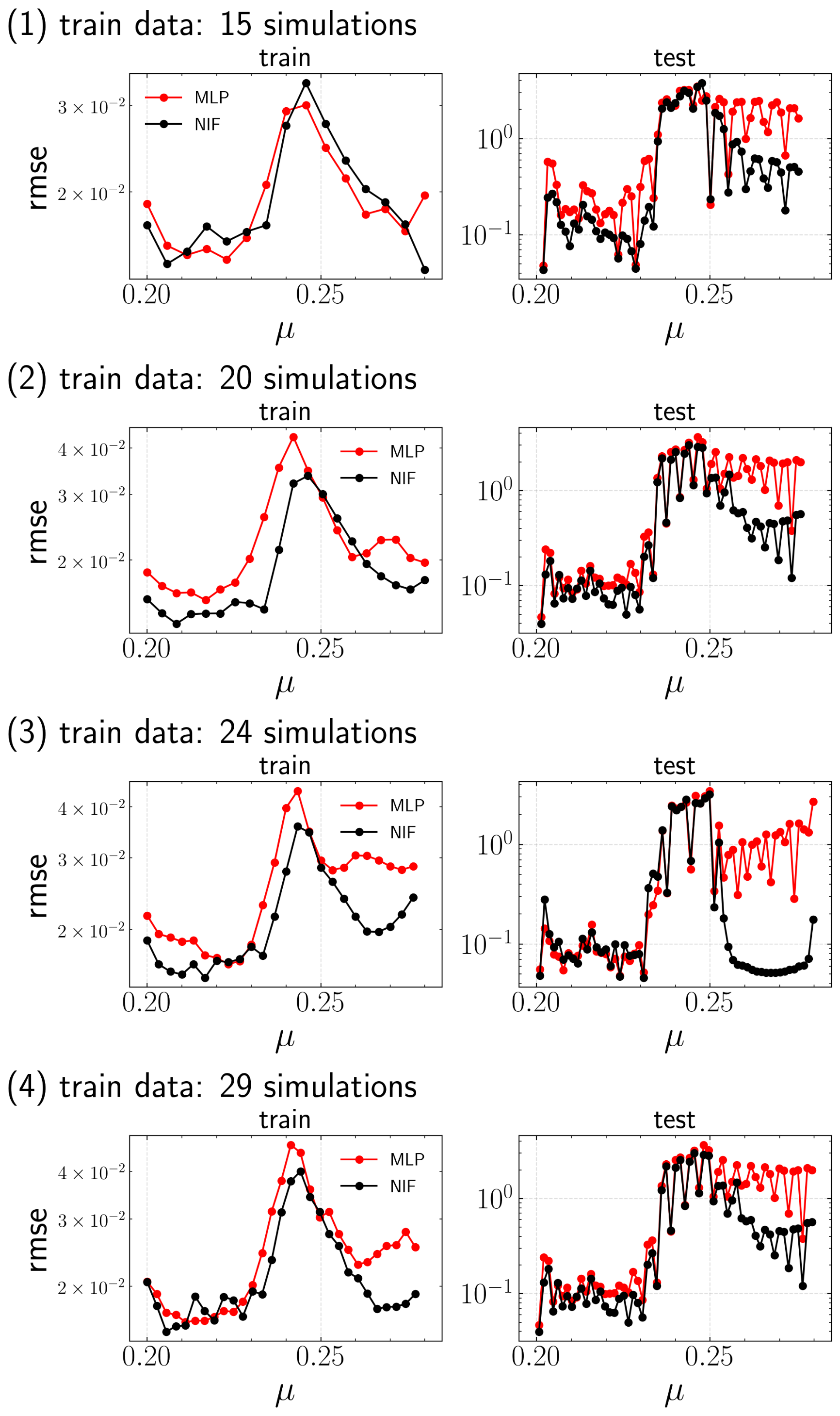}
\caption{Comparison between NIF-Swish and MLP-Swish in terms of RMSE on the dataset of 1D parametric KS system with $0.2<\mu<0.28$. Training data contains different numbers of simulations from 15 to 29.}
\label{fig:ks_pm_apdx_change_data}
\end{figure}

Here we consider further changing the size of training data to verify if NIF-Swish still performs better than MLP-Swish. As shown in \cref{fig:ks_pm_apdx_change_data}, we prepare another three set of datasets with 15, 24, and 29 samples of parameters, which correspond to simulations of K-S system at those parameters.\Cref{fig:ks_pm_apdx_change_data} shows the same phenomenon as before that NIF (Swish) generalizes better than MLP (Swish) when $\mu > 0.25$. 
\begin{table}[t]
\centering
\scriptsize
\caption{Comparison between standard MLP and NIF in \cref{sec:datafit} for surrogate modeling of 1D parametric PDE. RMSE below is averaged over all parameter $\mu$.}
\label{tab:apdx-network-structure}
\vspace{1em}
\begin{tabular}{cccc}
\textbf{Model}   & \textbf{ShapeNet}/\textbf{Vanilla MLP} & \textbf{ParameterNet} & \textbf{Number of training parameters}\\ 
\midrule
NIF (Swish)-1 & 1-30-30-30-1 & 2-30-30-2-1951  & 6,935 \\
MLP (Swish)-1 & 3-58-58-58-1 &  & 7,135 \\
NIF (Swish)-2 & 1-38-38-38-1 & 2-29-29-2-3079  & 10,254 \\
MLP (Swish)-2 & 3-70-70-70-1 &  & 10,291 \\
NIF (Swish)-3 & 1-56-56-56-1 &2-30-30-2-6553  & 20,741 \\
MLP (Swish)-3 & 3-100-100-100-1 &  & 20,701 \\
NIF (Swish)-4 & 1-60-60-60-1 & 2-47-47-2-7501  & 24,996 \\
MLP (Swish)-4 & 3-110-110-110-1 &  & 24,971 \\
NIF (Swish)-5 & 1-70-70-70-1 & 2-60-60-2-10151  & 34,415 \\
MLP (Swish)-5 & 3-130-130-130-1&  &34,711 \\
\bottomrule
\vspace{1em}
\end{tabular}
\end{table}


\begin{figure}[t]
\vspace{-.2in}
\centering
\includegraphics[width=0.63\textwidth]{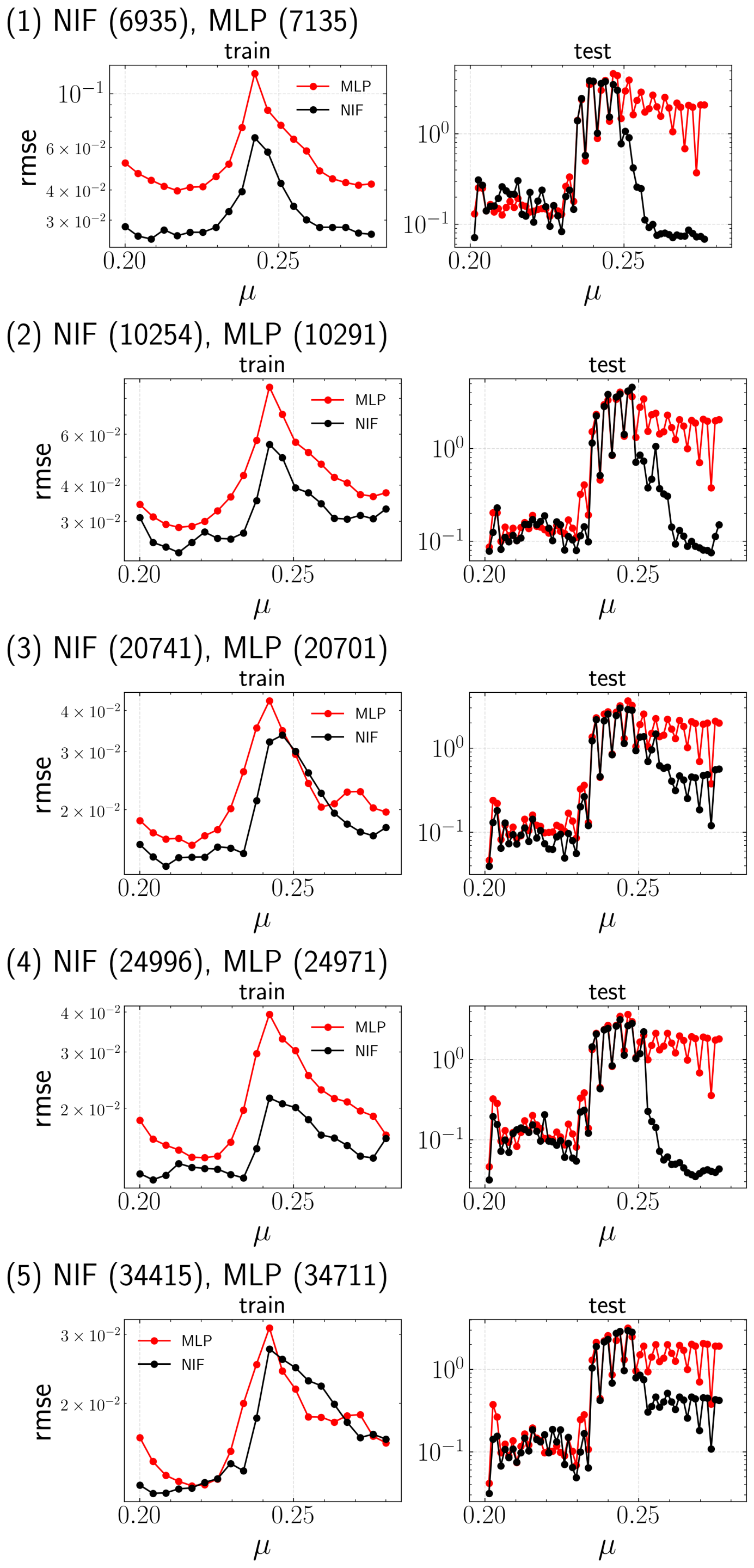}
\vspace{-.1in}
\caption{Comparison between NIF-Swish and MLP-Swish in terms of RMSE on the dataset of 1D parametric KS system with $0.2<\mu<0.28$. Training data contains 20 simulations but Model parameters vary from 7k to 34k.}
\label{fig:ks_pm_apdx_change_model}
\end{figure}

Finally, we consider verifying the above phenomenon by changing the number of trainable parameters away from the baseline in \cref{sec:app_data-fit-1d-ks} as described in \cref{tab:apdx-network-structure}. Again, we still train for 4 converged runs and compute the average error. Still, \Cref{fig:ks_pm_apdx_change_model} shows the same phenomenon that explains the difference between NIF (Swish) and MLP (Swish).

\subsection{Comparison of autoencoders for 2D Rayleigh-Taylor instability}
\label{apdx:autoencoder}


\subsubsection{Detailed setup of convolutional autoencoder}
\label{apdx:cae-details}

The encoder first uses three consecutive stride-2 convolutional layer followed by batch normalization and Swish activations with the number of output channel as 16, 32, 64. Output of the above is followed by flatten layer and a linear dense layer with $r$ output units. Then such units are feed to the decoder, which first contains a linear dense layer followed by a reshape layer and two consecutive dilated transposed convolutional layer followed by batch normalization and Swish activation with the number of output channel as 32, 16. Finally, the last layer is a dilated transpose convolutional layer that maps to the same size as input snapshot. For the best results, we do not use any pooling layers throughout. 

\begin{table}[t]
\centering
\scriptsize
\caption{CAE architecture with input shape $128\times 256$.}
\label{tab:apdx-cae-structure}
\vspace{1em}
\begin{tabular}{ccc}
\textbf{Layer (type)}   & \textbf{Output Shape} & \textbf{Parameter Count} \\ 
\midrule
Convolution 2D & (None, 128, 64, 16) & 160 \\
Batch Normalization & (None, 128, 64, 16) & 64 \\
Swish Activation & (None, 128, 64, 16) & 0 \\
Convolution 2D & (None, 64, 32, 32) & 4640 \\
Batch Normalization & (None, 64, 32, 32) & 128 \\
Swish Activation & (None, 64, 32, 32) & 0 \\
Convolution 2D & (None, 32, 16, 64) & 18496 \\
Batch Normalization & (None, 32, 16, 64) & 256 \\
Swish Activation & (None, 128, 64, 16) & 0 \\
Flatten & (None, 32768)   & 0 \\
Dense & (None, 8)   & 262152 \\
Dense & (None, 32768)   & 294912 \\
Reshape & (None, 32, 16, 64)   & 0 \\
ConvolutionTranspose 2D & (None, 64, 32, 32) & 18464 \\
Batch Normalization & (None, 64, 32, 32) & 128 \\
Swish Activation & (None, 64, 32, 32) & 0 \\
ConvolutionTranspose 2D & (None, 128, 64, 16) & 4624 \\
Batch Normalization & (None, 128, 64, 16) & 64 \\
Swish Activation & (None, 128, 64, 16) & 0 \\
ConvolutionTranspose 2D & (None, 256, 128, 1) & 145\\ 
\bottomrule
\vspace{1em}
\end{tabular}
\end{table}

\subsubsection{Evolution of fitting, projection and total error}
\label{apdx:compare_time_evolution_err}

As mentioned before, we can quantitatively compare the performance of three methods of dimensionality reduction by projecting those outputs on a very fine mesh with 256$\times$512 resolution. Recall that the shape of POD output is 128$\times$256, and the shape of CAE output varyes from 32$\times$64, 64$\times$128, 128$\times$256. Hence, we use nearest-neighbor to obtain the projection onto 256$\times$512 using the \texttt{transform.resize} function in \texttt{scikit-image} package \citep{van2014scikit}. While for NIF, we simply perform spatial query at those coordinates of cell nodes, which are $x_i=2.5\times 10^{-7} + (i-1) \times 1.96078333 \times 10^{-3}$, $y_j=5 \times 10^{-7} + (j-1) \times  1.95694618 \times 10^{-3}$, for $i=1,\ldots,256$, $j=1,\ldots,512$. This can be generated in Python with \texttt{numpy.linspace(2.5e-7, 0.5, 256, endpoint=True)} and \texttt{numpy.linspace(5e-7, 1, 512, endpoint=True)}.

Unless highlighted, the following data fields are projected onto 256$\times$512 mesh and become 2D arrays varying with time $t$. Value of the field at each coordinate $(x_i,y_j)$ and time $t$ can be indexed by $(i,j; t)$.  
\begin{itemize}
\item $u_{\textrm{true}}$: ground true data from adaptive mesh,
\item $u_{\textrm{true}, 32\times 64}$: ground true data sampled on the $32\times 64$ Cartesian mesh,
\item $u_{\textrm{true}, 64\times 128}$: ground true data sampled on the $64\times 128$ Cartesian mesh,
\item $u_{\textrm{true}, 128\times 256}$: ground true data sampled on the $128\times 256$ Cartesian mesh,
\item $u_{\textrm{POD}, 128\times 256}$: output prediction from POD on the $128\times 256$ Cartesian mesh,
\item $u_{\textrm{CAE}, 32\times 64}$: output prediction from CAE on the $32\times 64$ Cartesian mesh,
\item $u_{\textrm{CAE}, 64\times 128}$: output prediction from CAE on the $64\times 128$ Cartesian mesh,
\item $u_{\textrm{CAE}, 128\times 256}$: output prediction from CAE on the $128\times 256$ Cartesian mesh,
\item $u_{\textrm{NIF}}$ as the output prediction from NIF evaluated on the 256$\times$512 mesh. 

\end{itemize}

Without loss of generality, let's take the CAE with training data on the $32\times 64$ mesh for example. Notice that 
\begin{equation}
\underbrace{u_{\textrm{true}} - u_{\textrm{CAE}, 32\times 64}}_{\textrm{total difference}} = \underbrace{u_{\textrm{true}} - u_{\textrm{true}, 32\times 64}}_{\textrm{projection difference}} + \underbrace{u_{\textrm{true}, 32\times 64} - u_{\textrm{CAE}, 32\times 64}}_{\textrm{fitting difference}}.
\end{equation}
Hence, we define the following three error metrics at each time $t$:
\begin{itemize}
\item \emph{fitting error}: spatially averaged mean square of fitting difference, 
\begin{equation}
\varepsilon_{\textrm{fitting}}^{\textrm{CAE}, 32\times 64} (t) = \frac{1}{256\times 512} \sum_{i=1}^{256} \sum_{j=1}^{512} (u_{\textrm{true}, 32\times 64}(i,j; t) - u_{\textrm{CAE}, 32\times 64}(i,j; t))^2.
\end{equation}

\item \emph{projection error}: spatially averaged square of project difference, 
\begin{equation}
\varepsilon_{\textrm{projection}}^{\textrm{CAE}, 32\times 64} (t)= \frac{1}{256\times 512} \sum_{i=1}^{256} \sum_{j=1}^{512} (u_{\textrm{true}}(i,j; t) - u_{\textrm{true}, 32\times 64}(i,j; t))^2.
\end{equation}

\item \emph{total error}: spatially averaged square of total difference
\begin{equation}
\varepsilon_{\textrm{total}}^{\textrm{CAE}, 32\times 64} (t) = \frac{1}{256\times 512} \sum_{i=1}^{256} \sum_{j=1}^{512} (u_{\textrm{true}}(i,j; t) - u_{\textrm{CAE}, 32\times 64}(i,j; t))^2.
\end{equation}

\end{itemize}

\begin{figure}[t]
\centering
\includegraphics[width=0.9\textwidth]{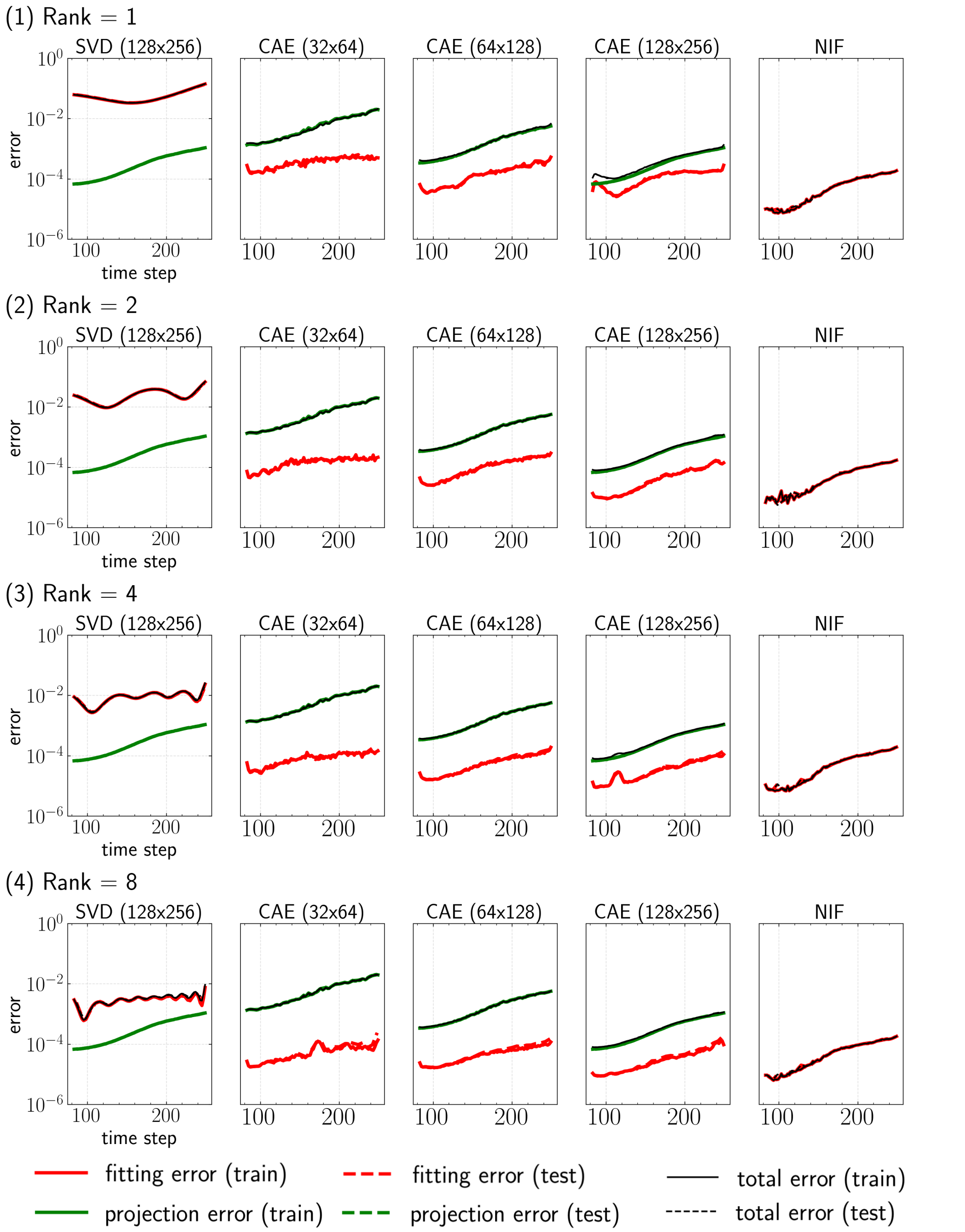}
\caption{Evolution of fitting, projection and total error for all five dimensionality reductions methods on learning low dimensional latent subspace with $r=2$ (top), $r=4$ (middle), $r=8$ (bottom).}
\label{fig:autoencoder_error_compare}
\end{figure}

We can define the same metrics for other models. Evolution of the above three error metrics for all the models on training and testing time steps are displayed in the first row of \cref{fig:autoencoder_error_compare}. Fitting error (red) contributes the most in POD while projection error is more than two orders of magnitude smaller. This implies the lack of expressiveness in POD leads to its inferior performance. In contrast, projection error (green) contributes most in CAE while fitting error remains unchanged with varying resolution of projection from 32$\times 64$ to 128$\times$256. This indicates that CAE is mainly restricted by the projection error introduced during preprocessing in this example. Meanwhile, NIF learns the latent representation directly from the raw data on adaptive mesh. Therefore, fitting error is the same as total error. Moreover, we observe that projection error grows \textit{near-exponentially} in time. This is because of the nature of R-T instability that energy from large scale structures transfer to small-scale as time goes. Such small-scale vortex generates even further smaller vortex. Eventually, the Cartesian grid becomes unable to resolve the flow, which ends up with a significant amount of projection error. As shown in the rest of \cref{fig:autoencoder_error_compare}, such phenomenon persists even changing rank $r$. The fact that the dimensionality of this R-T data is one is consistent with \cref{fig:autoencoder_error_compare} from which only POD methods improves when $r$ is increased.

\subsection{Sparse reconstruction on sea surface temperature}
\label{apdx:sparse_rec}

\begin{figure}[t]
\vspace{-.2in}
\centering
\includegraphics[width=.8\textwidth]{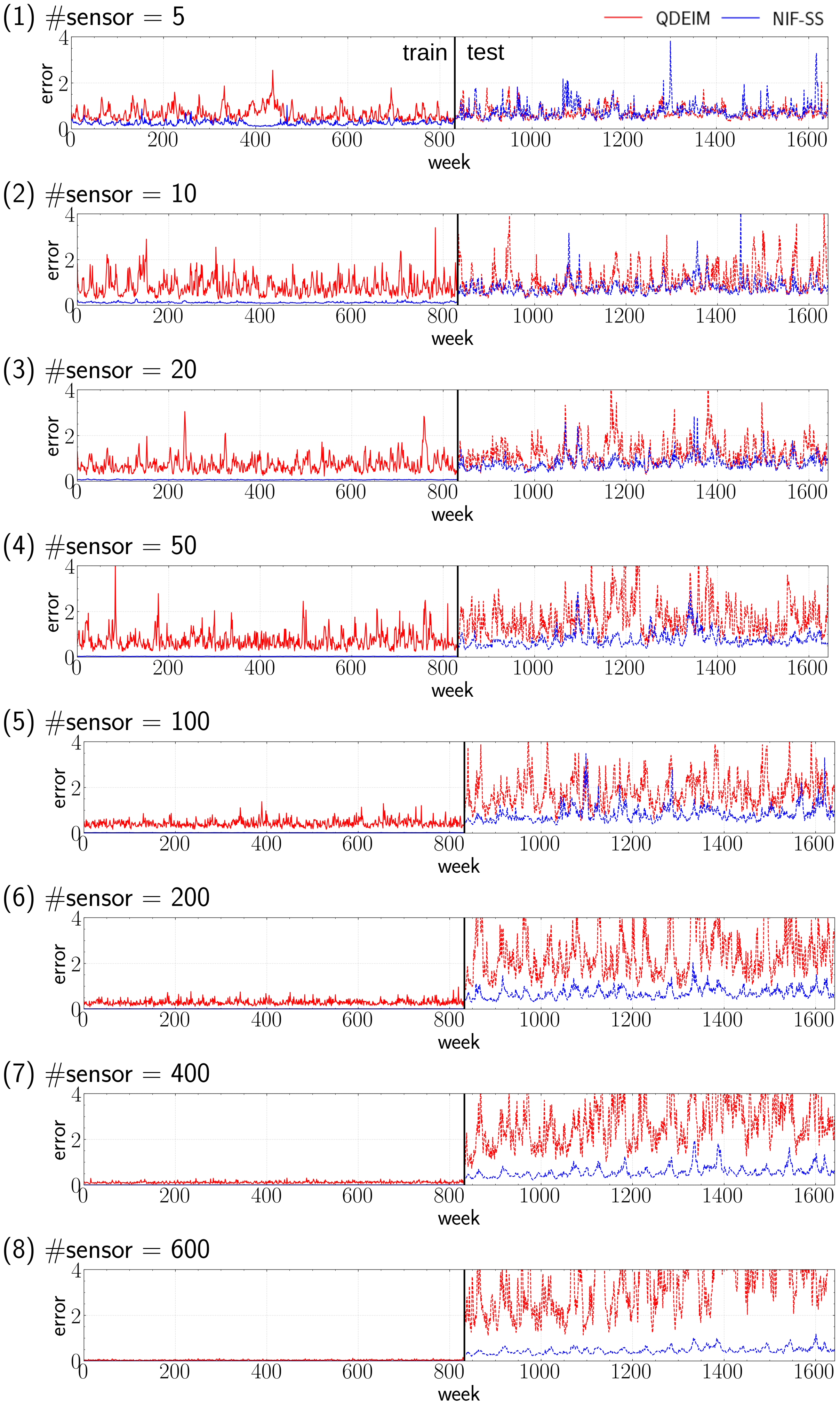}
\vspace{-.2in}
\caption{History of spatially mean-squared error of NIF-based framework and POD-QDEIM on sea surface temperature from 1990 to 2021  with different number of sensors $p$.}
\label{fig:sst_ss_err_history}
\end{figure}

Comparison on temporal variation of spatially mean-squared error between our framework and POD-QDEIM~\citep{manohar2018data} with different number of sensors $p$ is shown in \cref{fig:sst_ss_err_history}. We can visually confirm the error variation is much higher in POD-QDEIM compared to our models. This in turn means the model prediction from POD-QDEIM should be accompanied with larger uncertainties. As one increases the number of sensors, training error of both models decay drastically. Meanwhile, testing error of POD-QDEIM increases and that of NIF-based framework visually stays around the same level. 

\begin{figure}[t]
\vspace{-.3in}
\centering
\includegraphics[width=0.75\textwidth]{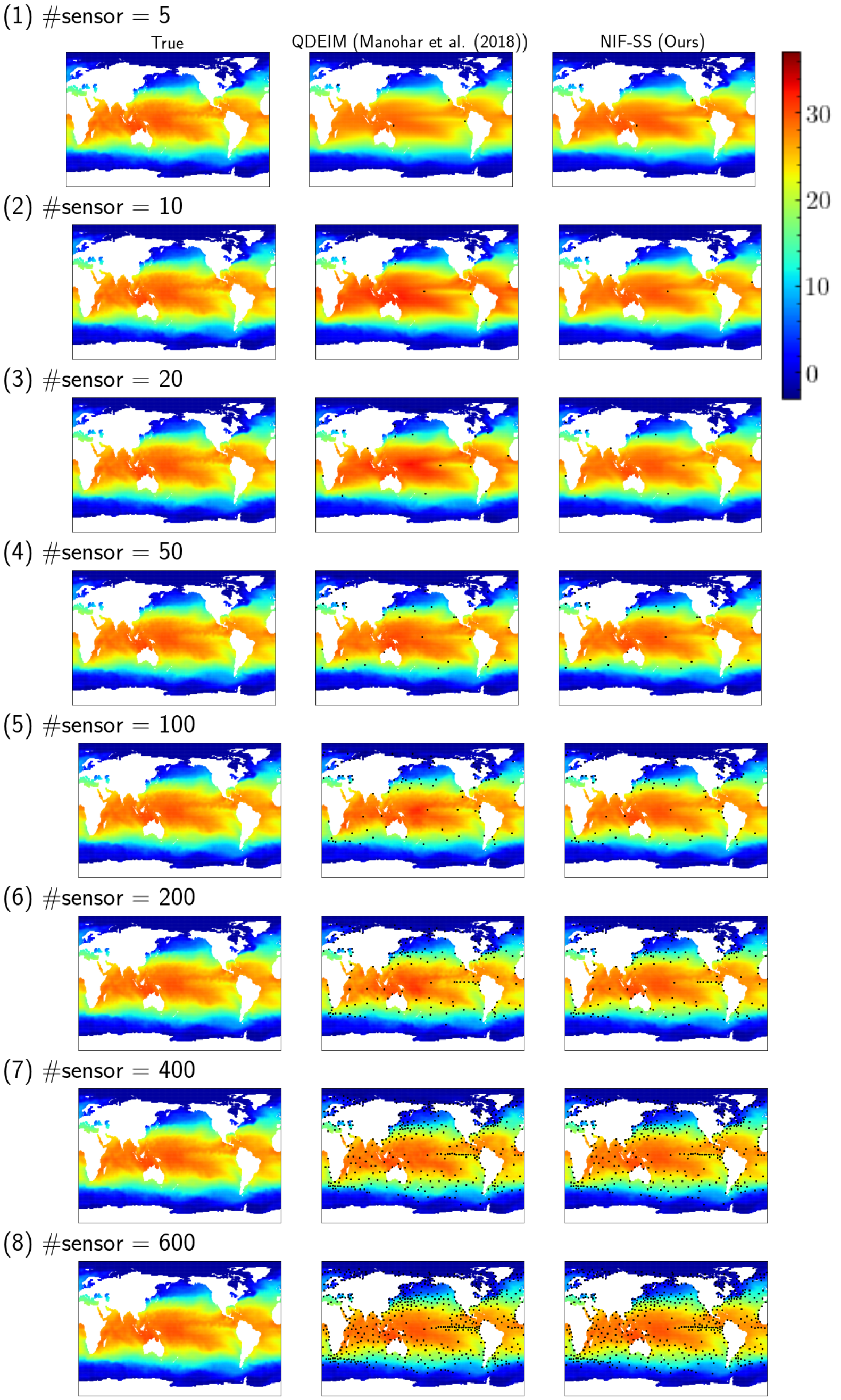}
\vspace{-.1in}
\caption{Sea surface temperature contour of the first week of 1990 from ground true (left column), POD-QDEIM (middle column) and our framework (right column)  with varying number of sensors. Black dots indicates optimized sensor location from POD-QDEIM.}
\label{fig:sst_ss_train_contour_temperature}
\vspace{-.1in}
\end{figure}

Comparison on contours of sea surface temperature among ground true, our framework and POD-QDEIM for the very first week of 1990 (in training dataset) is displayed in \cref{fig:sst_ss_train_contour_temperature}. We can roughly see the inferior performance of POD-QDEIM comes from at least two sources: 
\begin{enumerate}
\item POD-QDEIM overshoots for the average temperature in the middle of the Pacific Ocean.
\item POD-QDEIM misses small scales structures while NIF-based framework captures them well.
\end{enumerate}
As the number of sensors increases, both models performs better and better on training data.

\begin{figure}[t]
\vspace{-.3in}
\centering
\includegraphics[width=0.75\textwidth]{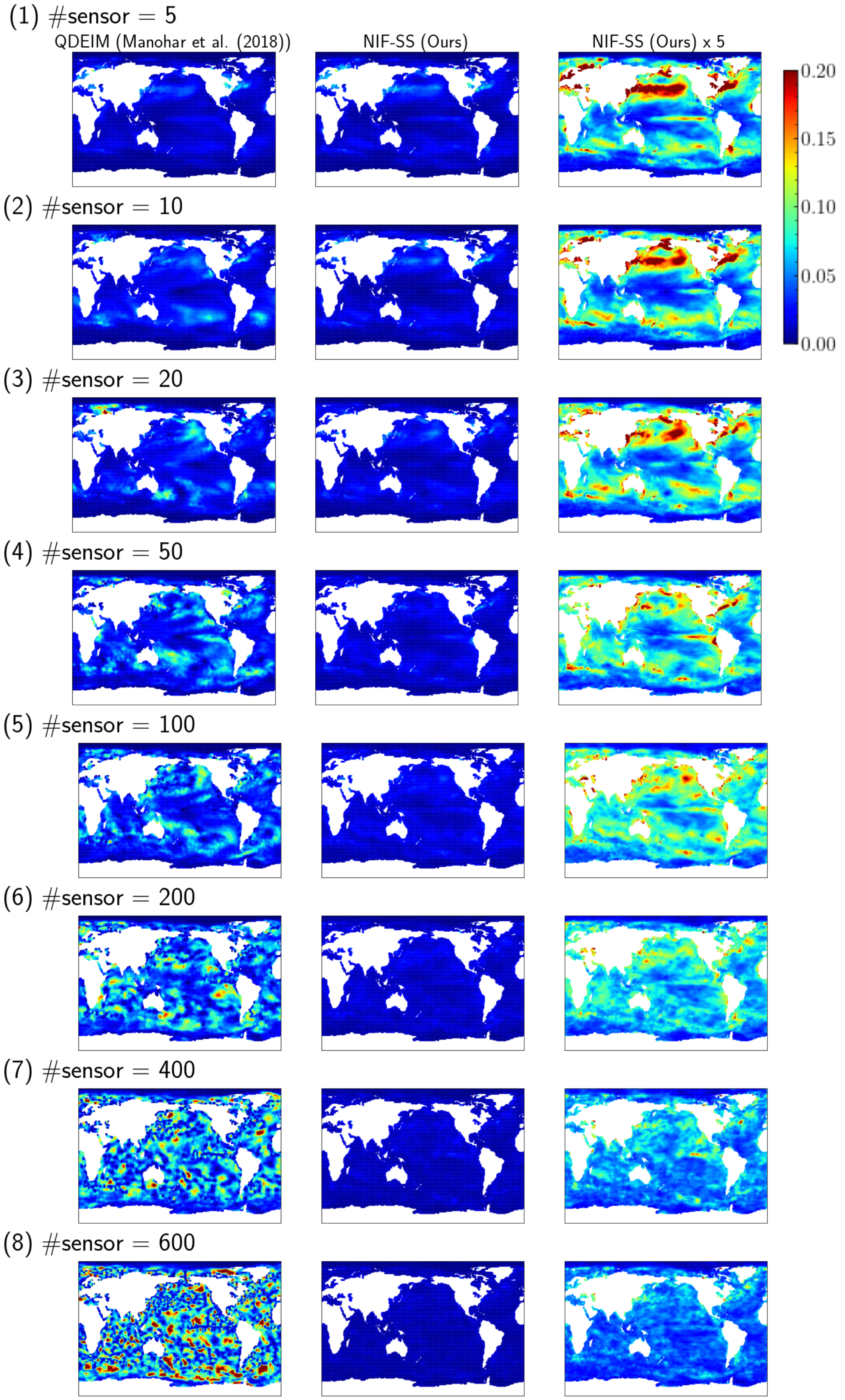}
\vspace{-.1in}
\caption{Spatial distribution of temporally mean squared error of NIF-based framework and POD-QDEIM with varying number of sensors. ``NIF-SS x 5'' means the error is magnified 5 times to increase visibility. }
\label{fig:sst_ss_contour_temperature_test_err}
\end{figure}

In order to further analyze the model generalization performance on sea surface temperature data, we compute the temporal average of squared error on unseen testing data (2006 to 2021) and plot its corresponding spatial distribution with varying number of sensors $p$ in \cref{fig:sst_ss_contour_temperature_test_err}. Evidently, except at 5 sensors where NIF-SS shows a similar testing error compared to POD-QDEIM, NIF-SS generalizes much better than POD-QDEIM regardless of the number of sensors $p$. As $p$ increases, the error distribution of both two frameworks tends to contain more small-scale patches. 

An inspection on error distribution from NIF-based framework shows interesting correlations between regions that are difficult to predict and ocean circulation patterns. Let's focus on the NIF-SS with error magnified 5 times (third column) in \cref{fig:sst_ss_contour_temperature_test_err}. For example, when $p=5$, we see the regions that are most difficult to predict by NIF-SS happen mostly at the location of \emph{Gulf stream}, \emph{North Pacific gyre} and \textit{Norwegian current}.

\section{Implementation of NIF with SIREN}

\label{apdx:nif_sine}

We adopt SIREN \citep{sitzmann2020implicit}, which is standard MLP with $\omega_0$-scaled sine activations, as our ShapeNet. In this work we take $\omega_0=30$ and find it to be sufficient for all cases. The initialization of SIREN is given in \cref{apdx_sec:init_siren}.  
Based on SIREN, we also borrow the ResNet-like structure \citep{lu2021compressive} to further improve training performance, which is presented in \cref{apdx_sec:resnet}. \Cref{apdx_sec:parameternet} describes how we connect ParameterNet with ShapeNet. Finally, practical advice on training NIF with SIREN is given in \cref{apdx_sec:advice_training}.

\subsection{Initialization of SIREN}
\label{apdx_sec:init_siren}

First of all, one should normalize each component of input for SIREN as $\mathcal{U}(-1,1)$ approximately. $\mathcal{U}(a,b)$ denotes uniform distribution on interval $[a,b]$, $a \le b \in \mathbb{R}$. For example, in \cref{apdx:siren_normalize}, we normalize one coordinate component $x$ as 
\begin{equation}
\label{apdx:siren_normalize}
\tilde{x} = \frac{x- (\min(x)+\max(x))/2}{(\max(x)-\min(x))/2}.
\end{equation}
For the output $u$, we choose the standard normalization  (zero mean and unit variance).

Second, SIREN requires a special initialization of weights and biases to achieve superior performance. Without loss of generality, consider a standard MLP equipped with $\omega_0$-scaled $\sin$ as activation functions and units structure as $n_i$-$n_h$-$n_h$-$n_h$-$n_o$. $n_i$/$n_o$ denotes the dimension of input/output layer. $n_h$ denotes the dimension of hidden layer. Next, we initialize weights and biases of input layer  component-wise in \cref{apdx:siren_1st_initialization} as 
\begin{equation}
\label{apdx:siren_1st_initialization}
\mathbf{W}_{1,(j,k)} \sim \mathcal{U}\left(-\frac{1}{n_i},\frac{1}{n_i}\right), \quad \bm{b}_{1,(j)} \sim \mathcal{U}\left( -\frac{1}{\sqrt{n_i}}, \frac{1}{\sqrt{n_i}} \right),
\end{equation}
where subscript $(j,k)$ denotes the $j$-th row $k$-th column component. We initialize all other layer weights and biases following \cref{apdx:siren_other_init}, 
\begin{equation}
\label{apdx:siren_other_init}
\mathbf{W}_{(j,k)} \sim \mathcal{U}\left(-\frac{\sqrt{6/n_h}}{\omega_0}, \frac{\sqrt{6/n_h}}{\omega_0} \right), \quad \bm{b}_{(j)} \sim \mathcal{U}\left( -\frac{1}{\sqrt{n_h}}, \frac{1}{\sqrt{n_h}} \right).
\end{equation}
Note that $\omega_0$-scaled sine activation is defined as $\sigma(x) = \sin(\omega_0 x )$.

\subsection{ResNet-like block}
\label{apdx_sec:resnet}

After first layer, we use a design of ResNet-like block \citep{lu2021compressive} to build ShapeNet. The configuration is displayed in \cref{fig:siren_nif_resnet}. Without loss of generality, denote $\eta_i$ as the input of $i$-th such block, we have 
\begin{align}
\zeta &= \sin(\omega_0\mathbf{W}_{i_1}\eta_i + \bm{b}_{i_1}),\\
\eta_{i+1} &= \frac{1}{2} \left(\eta_i + \sin(\omega_0 \mathbf{W}_{i_2} \zeta  + \bm{b}_{i_2} )\right),
\end{align}
where $\eta_{i+1}$ is the input for the next $i+1$-th block.

\subsection{Building ParameterNet}
\label{apdx_sec:parameternet}

The final step is to connect the output of ParameterNet to ShapeNet. Note that now only ParameterNet contains undetermined parameters while the parameters in ShapeNet are subject to the output of ParameterNet. Therefore, we only need to carefully design the initialization of last layer weights and biases of ParameterNet in order to be consistent with the requirement in \cref{apdx_sec:init_siren}. Recall that the network structure of NIF is a special case of hypernetworks of MLP that focuses on decoupling spatial complexity away from other factors. We take the initialization scheme of hypernetworks in SIREN \citep{sitzmann2020implicit}. The idea is to simply multiply the last layer initial weights by a small factor, e.g., $10^{-2}$, while keeping the last layer initial biases to match the distribution in \cref{apdx_sec:init_siren}.

\subsection{Practical advice for training}
\label{apdx_sec:advice_training}

We empirically found successful and stable training NIF with SIREN using Adam optimizer requires 1) \textit{small} learning rate typically around $10^{-4}$ to $10^{-5}$, 2) \textit{large} batch size, which often takes to be filling up the whole GPU memory. When debugging, it is recommended to monitor training error at every 5 epoch and plot the prediction at the same time. If one found training error, e.g., mean-square error, stuck at relatively high in the first 40 epoch and the prediction is completely useless, then one should further decrease the learning rate or increasing the batch size. We found NIF with SIREN also tend to fit large scale structure first then small scales. 

However, it is not always possible to increase the batch size given limited GPU memory especially in the fully 3D case. We found that if we fully parameterize the weights and biases of ShapeNet (as oppose to only parameterizing the last layer in \cref{sec:linspace}), the memory footprints of NIF becomes relatively large due to the tensor multiplication with \texttt{einsum} in implementation. Therefore, the maximal number of epochs affordable can be limited, which can lead to too long training time or unstable training if even one set learning rate as low as $10^{-5}$ in the worst case. 

If one doesn't have access to good GPU resources, a more economic remedy on a single GPU is to use \textit{small-batch} training, which is well-known to have better generalization compared to large-batch training \citep{masters2018revisiting}. However, here we are only looking for stable training of NIF with SIREN that uses small batch sizes to fit in a single GPU. We empirically found L4 optimizer \citep{rolinek2018l4} can have stable training performance in the \textit{small} batch regime compared to Adam for the same batch size. Empirically, we found L4 optimizer can reduce \textit{minimal batch size} required for stable training by at least 10 times compared to Adam optimizer, while only has a slight increase in training cost. Thus, one can increase the capacity of NIF by 10 times without sacrificing much in the overall training time and performance. Note that such capacity can be crucial in training large-scale 3D fully turbulence dataset with limited GPU resources. 

For all the results shown in this paper, we have used Nvidia Tesla P100 (16 GB), Nvidia GeForce RTX 2080 GPU (12 GB), and Nvidia A6000 GPU (48 GB). If available, the simplest remedy is to use data-parallelism with multiple GPUs.  We have implemented data-parallel capability for NIF in our Github repo (\texttt{https://github.com/pswpswpsw/nif}) on multiple GPUs and it scales well. We leave the complete study on distributed learning with NIF for future work.

\vskip 0.2in
\bibliography{sample}

\end{document}